\documentclass[twoside]{article}

\usepackage[accepted]{aistats2022}
\usepackage[round]{natbib}

\usepackage[latin9]{inputenc}
\usepackage{amsmath}
\usepackage{amssymb}
\usepackage{graphicx}
\usepackage[unicode=true,
 bookmarks=false,
 breaklinks=true,pdfborder={0 0 1},backref=section,colorlinks=false]
 {hyperref}

\makeatletter

\pdfpageheight\paperheight
\pdfpagewidth\paperwidth

\providecommand{\tabularnewline}{\\}

\usepackage{epsfig}
\usepackage{graphicx}

\@ifundefined{showcaptionsetup}{}{%
 \PassOptionsToPackage{caption=false}{subfig}}
\usepackage{subfig}
\makeatother

\begin{document}

%

%

\twocolumn[

\aistatstitle{Understanding and Achieving Efficient Robustness with \\
Adversarial Supervised Contrastive Learning}

\aistatsauthor{ Anh Bui \And Trung Le \\ \And  He Zhao}

\aistatsaddress{ Monash University \And  Monash University \And  Monash University} 

\aistatsauthor{Paul Montague \And Seyit Camtepe }

\aistatsaddress{Defence Science and Technology Group, Australia \And CSIRO Data61, Australia } 

\aistatsauthor{Dinh Phung }

\aistatsaddress{Monash University } 

]

\global\long\def\sidenote#1{\marginpar{\small\emph{{\color{Medium}#1}}}}%

\global\long\def\se{\hat{\text{se}}}%
\global\long\def\interior{\text{int}}%
\global\long\def\boundary{\text{bd}}%
\global\long\def\ML{\textsf{ML}}%
\global\long\def\GML{\mathsf{GML}}%
\global\long\def\HMM{\mathsf{HMM}}%
\global\long\def\support{\text{supp}}%
\global\long\def\new{\text{*}}%
\global\long\def\stir{\text{Stirl}}%
\global\long\def\expect{\mathbb{E}}%
\global\long\def\dist{d}%
\global\long\def\HX{\entro\left(X\right)}%
\global\long\def\entropyX{\HX}%
\global\long\def\HY{\entro\left(Y\right)}%
\global\long\def\entropyY{\HY}%
\global\long\def\HXY{\entro\left(X,Y\right)}%
\global\long\def\entropyXY{\HXY}%
\global\long\def\mutualXY{\mutual\left(X;Y\right)}%
\global\long\def\mutinfoXY{\mutualXY}%
\global\long\def\goto{\rightarrow}%
\global\long\def\asgoto{\stackrel{a.s.}{\longrightarrow}}%
\global\long\def\pgoto{\stackrel{p}{\longrightarrow}}%
\global\long\def\dgoto{\stackrel{d}{\longrightarrow}}%
\global\long\def\lik{\mathcal{L}}%
\global\long\def\logll{\mathit{l}}%
\global\long\def\bigcdot{\raisebox{-0.5ex}{\scalebox{1.5}{\ensuremath{\cdot}}}}%
\global\long\def\sig{\textrm{sig}}%
\global\long\def\likelihood{\mathcal{L}}%
\global\long\def\vectorize#1{\mathbf{#1}}%

\global\long\def\vt#1{\mathbf{#1}}%
\global\long\def\gvt#1{\boldsymbol{#1}}%
\global\long\def\idp{\ \bot\negthickspace\negthickspace\bot\ }%
\global\long\def\cdp{\idp}%
\global\long\def\das{}%
\global\long\def\id{\mathbb{I}}%
\global\long\def\idarg#1#2{\id\left\{  #1,#2\right\}  }%
\global\long\def\iid{\stackrel{\text{iid}}{\sim}}%
\global\long\def\bzero{\vt 0}%
\global\long\def\bone{\mathbf{1}}%
\global\long\def\a{\mathrm{a}}%
\global\long\def\ba{\mathbf{a}}%
\global\long\def\b{\mathrm{b}}%
\global\long\def\bb{\mathbf{b}}%
\global\long\def\B{\mathrm{B}}%
\global\long\def\boldm{\boldsymbol{m}}%
\global\long\def\c{\mathrm{c}}%
\global\long\def\C{\mathrm{C}}%
\global\long\def\d{\mathrm{d}}%
\global\long\def\D{\mathrm{D}}%
\global\long\def\N{\mathrm{N}}%
\global\long\def\h{\mathrm{h}}%
\global\long\def\H{\mathrm{H}}%
\global\long\def\bH{\mathbf{H}}%
\global\long\def\K{\mathrm{K}}%
\global\long\def\M{\mathrm{M}}%
\global\long\def\bff{\vt f}%
\global\long\def\bx{\mathbf{\mathbf{x}}}%

\global\long\def\bl{\boldsymbol{l}}%
\global\long\def\s{\mathrm{s}}%
\global\long\def\T{\mathrm{T}}%
\global\long\def\bu{\mathbf{u}}%
\global\long\def\v{\mathrm{v}}%
\global\long\def\bv{\mathbf{v}}%
\global\long\def\bo{\boldsymbol{o}}%
\global\long\def\bh{\mathbf{h}}%
\global\long\def\bs{\boldsymbol{s}}%
\global\long\def\x{\mathrm{x}}%
\global\long\def\bx{\mathbf{x}}%
\global\long\def\bz{\mathbf{z}}%
\global\long\def\hbz{\hat{\bz}}%
\global\long\def\z{\mathrm{z}}%
\global\long\def\y{\mathrm{y}}%
\global\long\def\bxnew{\boldsymbol{y}}%
\global\long\def\bX{\boldsymbol{X}}%
\global\long\def\tbx{\tilde{\bx}}%
\global\long\def\by{\boldsymbol{y}}%
\global\long\def\bY{\boldsymbol{Y}}%
\global\long\def\bZ{\boldsymbol{Z}}%
\global\long\def\bU{\boldsymbol{U}}%
\global\long\def\bn{\boldsymbol{n}}%
\global\long\def\bV{\boldsymbol{V}}%
\global\long\def\bI{\boldsymbol{I}}%
\global\long\def\J{\mathrm{J}}%
\global\long\def\bJ{\mathbf{J}}%
\global\long\def\w{\mathrm{w}}%
\global\long\def\bw{\vt w}%
\global\long\def\bW{\mathbf{W}}%
\global\long\def\balpha{\gvt{\alpha}}%
\global\long\def\bdelta{\boldsymbol{\delta}}%
\global\long\def\bsigma{\gvt{\sigma}}%
\global\long\def\bbeta{\gvt{\beta}}%
\global\long\def\bmu{\gvt{\mu}}%
\global\long\def\btheta{\boldsymbol{\theta}}%
\global\long\def\blambda{\boldsymbol{\lambda}}%
\global\long\def\bgamma{\boldsymbol{\gamma}}%
\global\long\def\bpsi{\boldsymbol{\psi}}%
\global\long\def\bphi{\boldsymbol{\phi}}%
\global\long\def\bpi{\boldsymbol{\pi}}%
\global\long\def\bomega{\boldsymbol{\omega}}%
\global\long\def\bepsilon{\boldsymbol{\epsilon}}%
\global\long\def\btau{\boldsymbol{\tau}}%
\global\long\def\bxi{\boldsymbol{\xi}}%
\global\long\def\realset{\mathbb{R}}%
\global\long\def\realn{\realset^{n}}%
\global\long\def\integerset{\mathbb{Z}}%
\global\long\def\natset{\integerset}%
\global\long\def\integer{\integerset}%

\global\long\def\natn{\natset^{n}}%
\global\long\def\rational{\mathbb{Q}}%
\global\long\def\rationaln{\rational^{n}}%
\global\long\def\complexset{\mathbb{C}}%
\global\long\def\comp{\complexset}%

\global\long\def\compl#1{#1^{\text{c}}}%
\global\long\def\and{\cap}%
\global\long\def\compn{\comp^{n}}%
\global\long\def\comb#1#2{\left({#1\atop #2}\right) }%
\global\long\def\nchoosek#1#2{\left({#1\atop #2}\right)}%
\global\long\def\param{\vt w}%
\global\long\def\Param{\Theta}%
\global\long\def\meanparam{\gvt{\mu}}%
\global\long\def\Meanparam{\mathcal{M}}%
\global\long\def\meanmap{\mathbf{m}}%
\global\long\def\logpart{A}%
\global\long\def\simplex{\Delta}%
\global\long\def\simplexn{\simplex^{n}}%
\global\long\def\dirproc{\text{DP}}%
\global\long\def\ggproc{\text{GG}}%
\global\long\def\DP{\text{DP}}%
\global\long\def\ndp{\text{nDP}}%
\global\long\def\hdp{\text{HDP}}%
\global\long\def\gempdf{\text{GEM}}%
\global\long\def\rfs{\text{RFS}}%
\global\long\def\bernrfs{\text{BernoulliRFS}}%
\global\long\def\poissrfs{\text{PoissonRFS}}%
\global\long\def\grad{\gradient}%
\global\long\def\gradient{\nabla}%
\global\long\def\partdev#1#2{\partialdev{#1}{#2}}%
\global\long\def\partialdev#1#2{\frac{\partial#1}{\partial#2}}%
\global\long\def\partddev#1#2{\partialdevdev{#1}{#2}}%
\global\long\def\partialdevdev#1#2{\frac{\partial^{2}#1}{\partial#2\partial#2^{\top}}}%
\global\long\def\closure{\text{cl}}%
\global\long\def\cpr#1#2{\Pr\left(#1\ |\ #2\right)}%
\global\long\def\var{\text{Var}}%
\global\long\def\Var#1{\text{Var}\left[#1\right]}%
\global\long\def\cov{\text{Cov}}%
\global\long\def\Cov#1{\cov\left[ #1 \right]}%
\global\long\def\COV#1#2{\underset{#2}{\cov}\left[ #1 \right]}%
\global\long\def\corr{\text{Corr}}%
\global\long\def\sst{\text{T}}%
\global\long\def\SST{\sst}%
\global\long\def\ess{\mathbb{E}}%

\global\long\def\Ess#1{\ess\left[#1\right]}%
\global\long\def\fisher{\mathcal{F}}%

\global\long\def\bfield{\mathcal{B}}%
\global\long\def\borel{\mathcal{B}}%
\global\long\def\bernpdf{\text{Bernoulli}}%
\global\long\def\betapdf{\text{Beta}}%
\global\long\def\dirpdf{\text{Dir}}%
\global\long\def\gammapdf{\text{Gamma}}%
\global\long\def\gaussden#1#2{\text{Normal}\left(#1, #2 \right) }%
\global\long\def\gauss{\mathbf{N}}%
\global\long\def\gausspdf#1#2#3{\text{Normal}\left( #1 \lcabra{#2, #3}\right) }%
\global\long\def\multpdf{\text{Mult}}%
\global\long\def\poiss{\text{Pois}}%
\global\long\def\poissonpdf{\text{Poisson}}%
\global\long\def\pgpdf{\text{PG}}%
\global\long\def\wshpdf{\text{Wish}}%
\global\long\def\iwshpdf{\text{InvWish}}%
\global\long\def\nwpdf{\text{NW}}%
\global\long\def\niwpdf{\text{NIW}}%
\global\long\def\studentpdf{\text{Student}}%
\global\long\def\unipdf{\text{Uni}}%
\global\long\def\transp#1{\transpose{#1}}%
\global\long\def\transpose#1{#1^{\mathsf{T}}}%
\global\long\def\mgt{\succ}%
\global\long\def\mge{\succeq}%
\global\long\def\idenmat{\mathbf{I}}%
\global\long\def\trace{\mathrm{tr}}%
\global\long\def\argmax#1{\underset{_{#1}}{\text{argmax}} }%
\global\long\def\argmin#1{\underset{_{#1}}{\text{argmin}\ } }%
\global\long\def\diag{\text{diag}}%
\global\long\def\norm{}%
\global\long\def\spn{\text{span}}%
\global\long\def\vtspace{\mathcal{V}}%
\global\long\def\field{\mathcal{F}}%
\global\long\def\ffield{\mathcal{F}}%
\global\long\def\inner#1#2{\left\langle #1,#2\right\rangle }%
\global\long\def\iprod#1#2{\inner{#1}{#2}}%
\global\long\def\dprod#1#2{#1 \cdot#2}%
\global\long\def\norm#1{\left\Vert #1\right\Vert }%
\global\long\def\entro{\mathbb{H}}%
\global\long\def\entropy{\mathbb{H}}%
\global\long\def\Entro#1{\entro\left[#1\right]}%
\global\long\def\Entropy#1{\Entro{#1}}%
\global\long\def\mutinfo{\mathbb{I}}%
\global\long\def\relH{\mathit{D}}%
\global\long\def\reldiv#1#2{\relH\left(#1||#2\right)}%
\global\long\def\KL{KL}%
\global\long\def\KLdiv#1#2{\KL\left(#1\parallel#2\right)}%
\global\long\def\KLdivergence#1#2{\KL\left(#1\ \parallel\ #2\right)}%
\global\long\def\crossH{\mathcal{C}}%
\global\long\def\crossentropy{\mathcal{C}}%
\global\long\def\crossHxy#1#2{\crossentropy\left(#1\parallel#2\right)}%
\global\long\def\breg{\text{BD}}%
\global\long\def\lcabra#1{\left|#1\right.}%
\global\long\def\lbra#1{\lcabra{#1}}%
\global\long\def\rcabra#1{\left.#1\right|}%
\global\long\def\rbra#1{\rcabra{#1}}%

\begin{abstract}
Contrastive learning (CL) has recently emerged as an effective approach to learning representation 
in a range of downstream tasks. Central to this approach is the selection of positive (similar) and 
negative (dissimilar) sets to provide the model the opportunity to `contrast' between data and class 
representation in the latent space. In this paper, we investigate CL for improving model robustness 
using adversarial samples. We first designed and performed a comprehensive study to understand how 
adversarial vulnerability behaves in the latent space. Based on this empirical evidence, we propose 
an effective and efficient supervised contrastive learning to achieve model robustness against 
adversarial attacks. Moreover, we propose a new sample selection strategy that optimizes the 
positive/negative sets by removing redundancy and improving correlation with the anchor. 
Extensive experiments show that our Adversarial Supervised Contrastive Learning (ASCL) approach achieves
comparable performance with the state-of-the-art defenses while significantly outperforms other CL-based defense methods 
by using only $42.8\%$ positives and $6.3\%$ negatives.
\end{abstract}

\section{Introduction\label{sec:intro}}

Recently, there has been a considerable research effort on adversarial
defense methods including \cite{akhtar2018threat,lecuyer2019certified,carlini2019evaluating,metzen2017detecting}
which aim to develop a robust Deep Neural Network against adversarial
attacks. Among them, the adversarial training methods (e.g, FGSM,
PGD adversarial training \citep{goodfellow2014explaining,madry2017towards}
and TRADES \citep{Zhang2019theoretically}) that utilize adversarial
examples as training data, have been one of the most effective series of approaches,
which truly boost the model robustness without the facing the problem
of obfuscated gradients \citep{athalye2018obfuscated}. In  adversarial
training, recently \cite{xie2019feature,bui2020improving} show
that reducing the divergence of the representations of images and
their adversarial examples in the latent space (e.g., the feature space
output from an intermediate layer of a classifier) can significantly
improve the robustness. For example, in \cite{bui2020improving},
the latent representations of images in the same class are pulled closer
together than those in different classes, which lead to a more compact
latent space and consequently, better robustness.

On the other hand, as proposed recently, contrastive learning (CL) has been an increasingly popular
and effective self-supervised representation learning approach \citep{chen2020simple,he2020momentum,khosla2020supervised}.
Specifically, CL learns representations of unlabeled data by choosing
an anchor $\bx_{i}$ and pulling the anchor and its positive samples
closer in latent space while pushing it away from many negative samples.
Intuitively, as the divergence in latent space is the focus of both AML and CL, it is natural to leverage CL to improve model robustness in adversarial training.
However, we in this paper demonstrate
that directly adopting CL into AML can hardly improve adversarial
robustness, indicating that a deeper understanding of the relationships
between the CL mechanism, latent space compactness, and adversarial
robustness is required.  Pursuing this comprehension, we give a detailed
study on the above aspects, and subsequently propose a new framework
for enhancing robustness using the principles of CL. Our paper provides
answers for three research questions: 

\begin{figure}
\begin{centering}
\includegraphics[width=1\columnwidth]{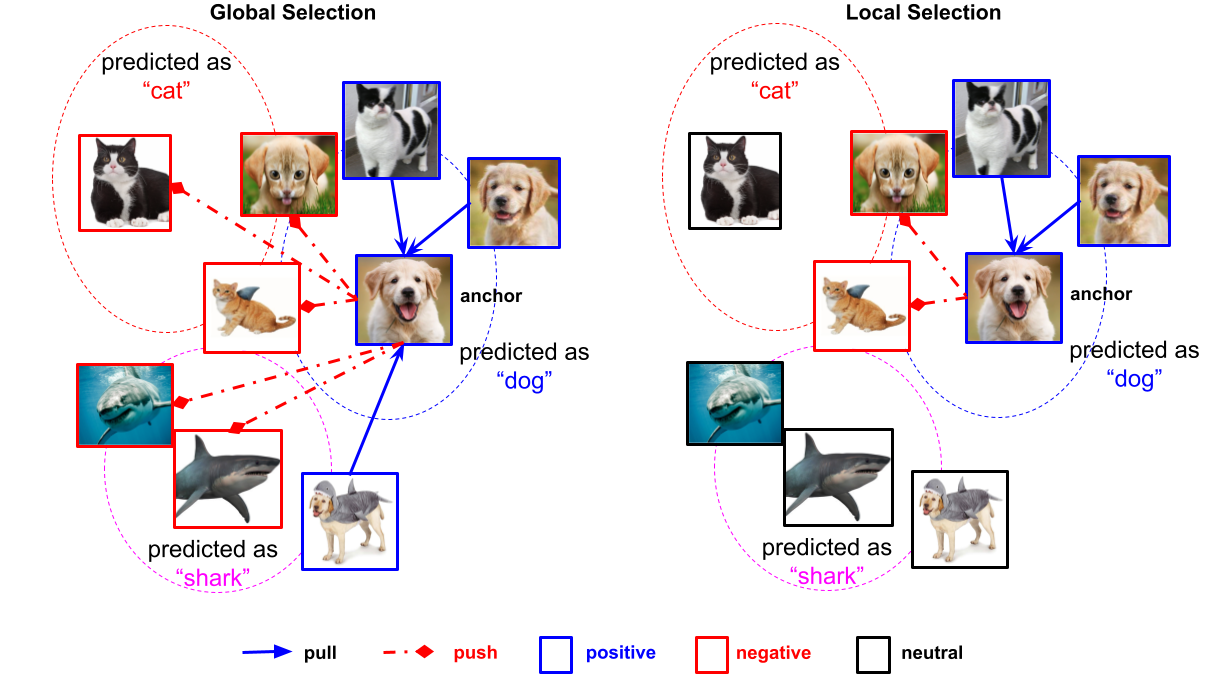}
\par\end{centering}
\caption{Illustration of ASCL with Global/Local Selection strategies in the
latent space. While Global Selection considers all other images in
the batch as either positives or negatives, Local Selection nominates
the most relevant samples to the anchor when operating contrastive
learning. The decision is based on the correlation between the true
labels and the predicted labels as in Table \ref{tab:3locals}.\label{fig:scl}}
\vspace{-7mm}
\end{figure}

\emph{(Q1) }\textbf{\emph{Why }}\emph{can CL help to improve the
adversarial robustness? }
To answer this question, we first introduce two kinds of divergences in the latent space:
the \textit{intra-class divergence} measured on benign
images and their adversarial examples of the same class and the \textit{inter-class
divergence} measured on those samples of different classes.
By comprehensively investigating the behavior
of divergence in latent space, our study shows that the robustness
of a model can be interpreted by the ratio between the intra- and inter-divergences: The
lower the ratio is, the more robustness can be achieved. These observations
motivate the idea that a robust model can be achieved by simultaneously
contrasting the intra-class divergence between images and their adversarial
examples with the inter-class divergence. We provide detailed analysis in Section \ref{sec:diver}.

\emph{(Q2) }\textbf{\emph{How }}\emph{to integrate CL with adversarial
training in the context of AML?} CL originally works with the case
where data labels are unavailable, which does not fit the context of AML for classifiers in the supervised setting.
The recent research of Supervised Contrastive Learning (SCL)
\citep{khosla2020supervised} extends CL by leveraging label information,
where the latent representations from the same class are pulled closer
together than those from different classes. While it might seem to be straightforward to apply SCL for AML, we show in this paper that it
is highly nontrivial to do so.
To this end, we propose \emph{Adversarial Supervised Contrastive Learning (ASCL)} to tackle this task by developing the following adaptions. Firstly, for
an anchor image, we use its adversarial images as the transformed/augmented
samples, which is different from the standard data augmentation techniques
used in conventional CL methods \citep{chen2020simple,khosla2020supervised}.
Secondly, we integrate SCL with adversarial training \citep{madry2017towards}
in addition to the clustering assumption \citep{chapelle2005semi},
to enforce compactness in latent space and subsequently improve
the adversarial robustness. 

\emph{(Q3) }\textbf{\emph{What}}\emph{ are the important factors for
the application of the ASCL framework in the context of AML?} One
of the key steps of CL/SCL is the selecting of positive and negative
samples for an anchor image. Although different approaches have been
proposed, most of them focus on natural images and can usually be ineffective
for AML. Specifically, in a data batch, CL and SCL consider the samples
that are not from the same instance or not in the same class of the
anchor image as its negative samples, which are hard splits between positive and negative sets, without taking
into account the correlation between a sample and the anchor image.
This can lead to too many true negative but useless samples which
are highly uncorrelated with the anchor in the latent space as illustrated
in Figure \ref{fig:scl}.  This issue aggravates with more diverse
data and in the AML context, making the original CL/SCL approaches
inapplicable. 
We therefore develop a novel series of strategies for selecting
positive and negative samples in our ASCL framework, which judiciously
picks the most relevant samples of the anchor that help to further improve
adversarial robustness.

By providing the answers to the above research questions,
we summarize our contributions in this paper as follows: 

\textbf{1)} We provide a comprehensive and insightful understanding of adversarial
robustness regarding the divergences in latent space, which sheds
light on adapting the CL principle to enhance robustness. 

\textbf{2)} We propose a novel Adversarial Supervised Contrastive Learning (ASCL)
framework, where the well-established contrastive learning mechanism
is leveraged to make the latent space of a classifier more compact,
leading to a more robust model against adversarial attacks. 

\textbf{3)} By analyzing the intrinsic characteristics of AML, we develop effective
strategies for selecting positive and negative samples more judiciously,
which are critical to making contrastive learning principle effective
in AML by using much less positives and negatives.

\textbf{4)} As shown in extensive experiments, our proposed framework is able
to significantly improve a classifier\textquoteright s robustness,
outperforming several adversarial training defense methods against strong attacks 
while achieving comparable performance with SOTA defenses in the RobustBench \citep{croce2020robustbench}.
\section{Analysis of Latent Space Divergence\label{sec:diver}}

By examining the question ``Why can CL help to improve the adversarial
robustness?'', we design experiments to show the connection of
adversarial robustness to the latent divergence of an anchor and
its contrastive samples.
\begin{center}
\begin{figure*}
\begin{centering}
\subfloat[Pairs of Absolute-DIVs with corresponding R-DIV. \label{fig:diver-compare-absolute}]{\begin{centering}
\includegraphics[width=0.28\textwidth]{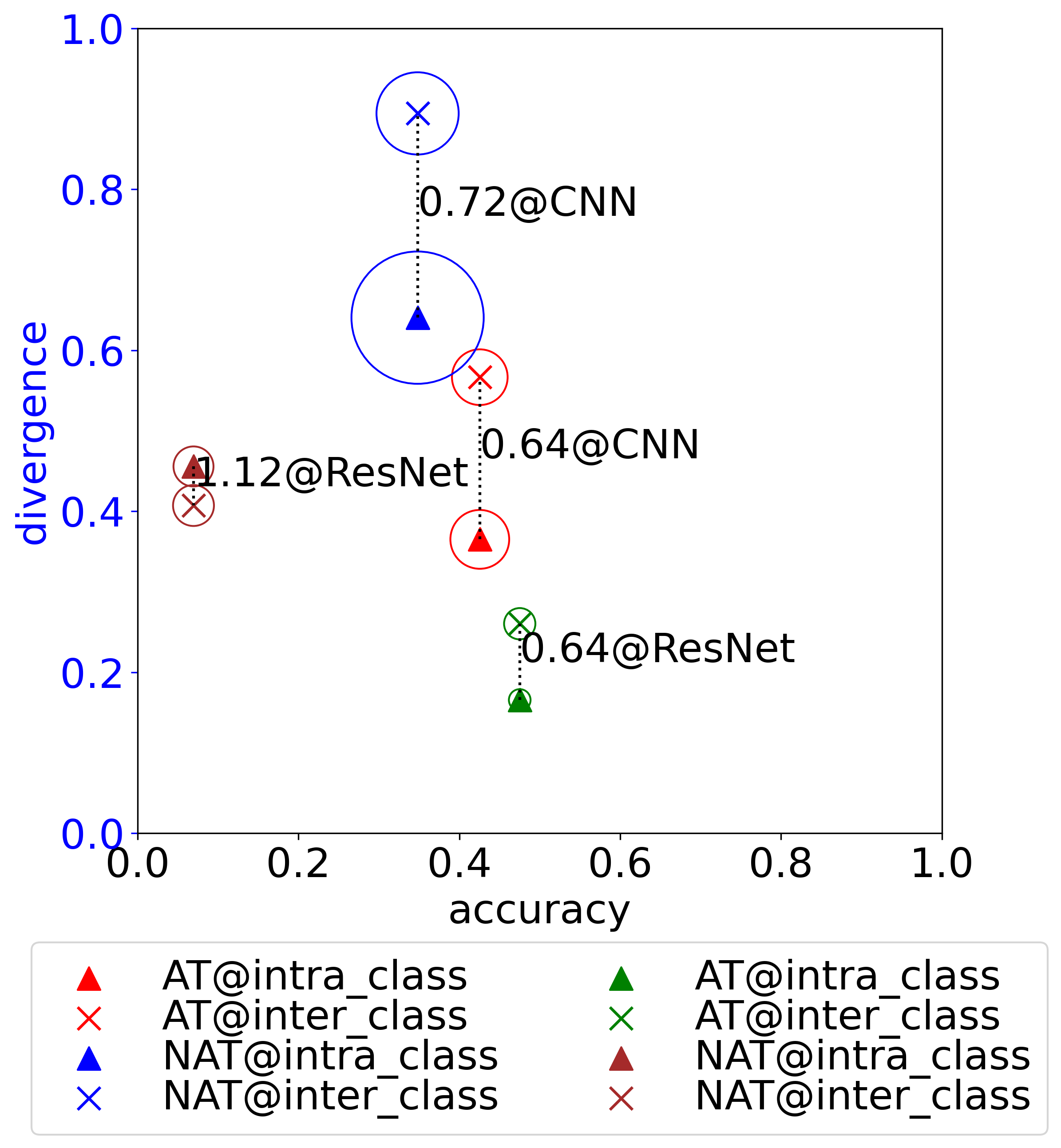}
\par\end{centering}
}\subfloat[R-DIV over the training progress with standard CNN model. \label{fig:diver-acc-progress}]{\begin{centering}
\includegraphics[width=0.33\textwidth]{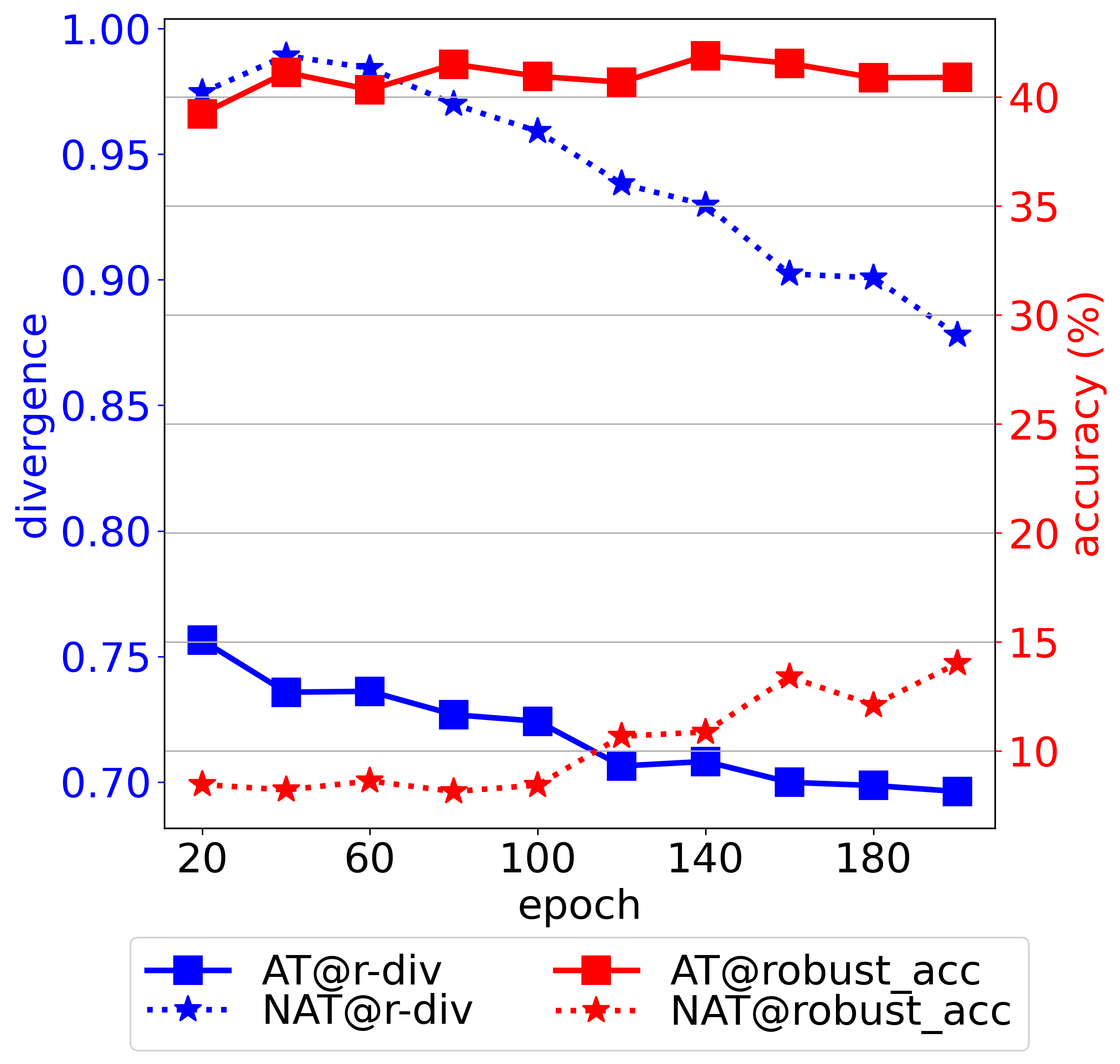}
\par\end{centering}
}\subfloat[R-DIV under different attack strengths. \label{fig:diver-acc}]{\begin{centering}
\includegraphics[width=0.3\textwidth]{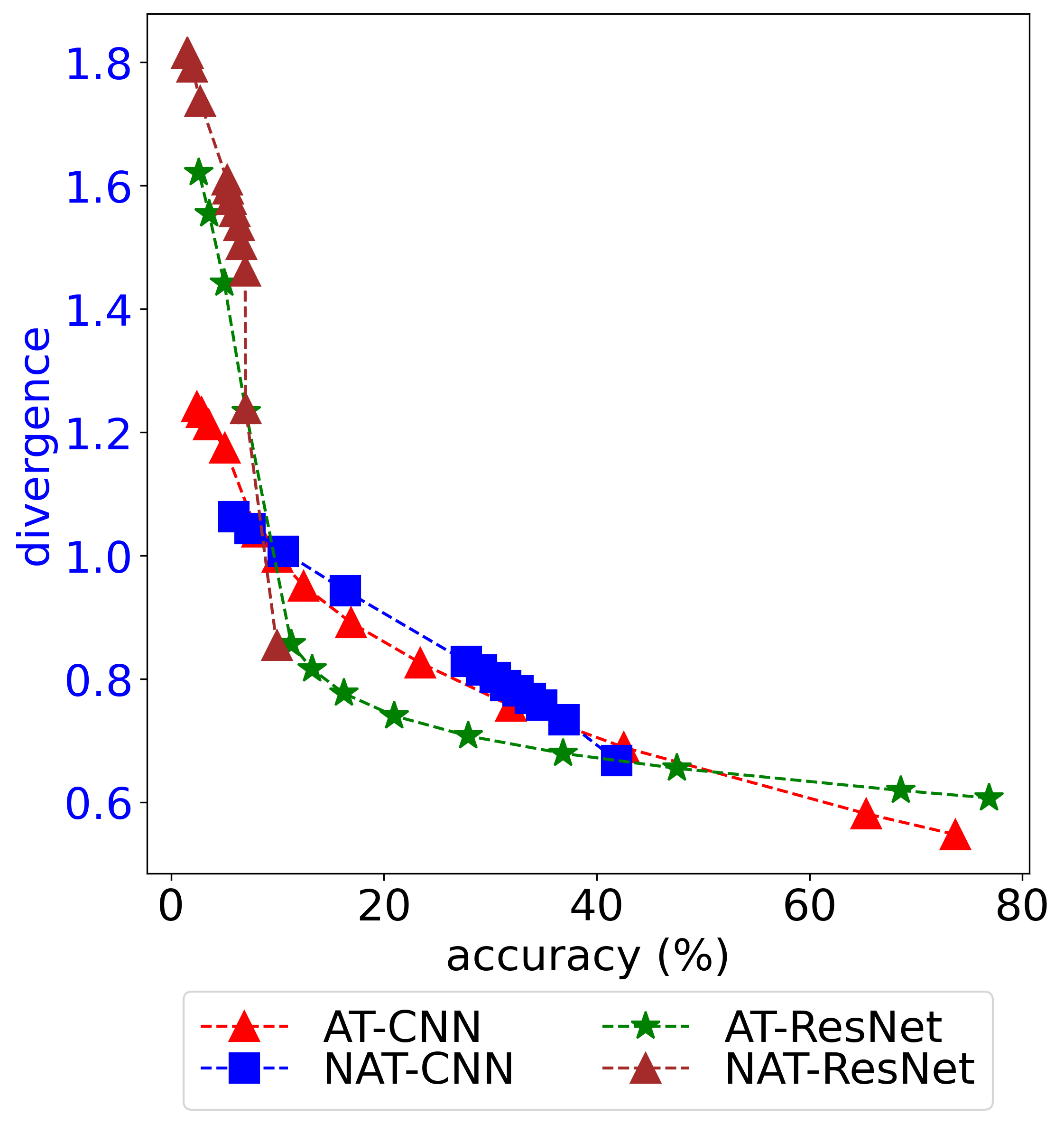}
\par\end{centering}
}
\par\end{centering}
\caption{Correlation between the\emph{ Relative intra-class} divergence (R-DIV)
and the robust accuracy on the CIFAR10 dataset. The variance of Absolute-DIV
in Figure \ref{fig:diver-compare-absolute} is scaled by 0.2 for
better visualization. Using PGD attack with $\epsilon=8/255,\eta=2/255$
with $k=10$ for training and $k=250$ for testing. 
\label{fig:divergence}}
\end{figure*}
\par\end{center}
\vspace{-10mm}
Let $\mathcal{\mathcal{B}}=\{\bx_{i},\by_{i}\}_{i=1}^{N}$ be a batch
of benign images where image $\bx_{i}$ is associated $\by_{i}$. Given an adversarial
transformation $\mathcal{T}$ from an adversary $\mathcal{A}$ (e.g.,
PGD attack in \citet{madry2017towards}), we consider two kinds of samples w.r.t. an anchor $\{\bx_{i},\by_{i}\}$:
the \emph{positive} set $\bX_{i}^{+}=\{\bx_{j},\bx_{j}^{\mathcal{T}}\mid j\neq i,\by_{j}=\by_{i}\}$
including benign examples and their counterparts in the same class
with the anchor and the \emph{negative} set $\bX_{i}^{-}=\{\bx_{j},\bx_{j}^{\mathcal{T}}\mid j\neq i,\by_{j}\neq\by_{i}\}$
including benign examples and their counterparts in different classes
with the anchor. 
We are interested in the latent representations of begin
and transformed images at a specific intermediate layer of the neural
net classifier $f$. Let us further denote those representations by
$\bz_{i}$ for benign images and $\bz_{i}^{\mathcal{T}}$ for adversarially transformed
images according $\mathcal{T}$. We define some types of divergences between benign images
and transformed images via transformation $\mathcal{T}$ at some intermediate
layers of $f$. 

(i)\emph{ }\textbf{\emph{Absolute}}\textbf{ intra-class divergence}:
$d_{a}^{+}=\frac{1}{N}\sum_{i=1}^{N}\frac{1}{\left|\bX_{i}^{+}\right|}\sum_{\bx_{j}\in\boldsymbol{X}_{i}^{+}}d(\bz_{i},\bz_{j})$
(i.e., evaluated based on the positive sets); and \textbf{\emph{absolute
}}\textbf{inter-class} \textbf{divergence}: $d_{a}^{-}=\frac{1}{N}\sum_{i=1}^{N}\frac{1}{\left|\bX_{i}^{-}\right|}\sum_{\bx_{j}\in\boldsymbol{X}_{i}^{-}}d(\bz_{i},\bz_{j})$
(i.e., evaluated based on the negative sets). Here we note that $d$
is cosine distance between two representations, and $\left|.\right|$
represents the cardinality of a set.

(ii) \textbf{\emph{Relative}}\textbf{ intra-class divergence} (R-DIV):
$d_{r}^{+}=\frac{d_{a}^{+}}{d_{a}^{-}}$; hence relative divergence
generally represents how large the magnitude of intra-class divergence
is relative to the inter-class divergence.

In Figure~\ref{fig:divergence}, we conduct an empirical study on the CIFAR-10 dataset to figure out the relationship
between R-DIV for adversarial examples and robust accuracy. The
findings and observations are very important for us to
devise our framework in the sequel. More specifically, we train a
CNN and a ResNet20 model in two modes: natural mode (NAT and cannot
defend at all) and adversarial training mode (AT and can defend quite
well). We observe how robust accuracy together with R-DIV vary with
training progress to draw conclusions. The detailed settings and
further demonstrations can be found in the supplementary material.
Some observations are drawn from our experiment:

(O1) \textbf{The robustness varies }\textbf{\emph{inversely}}\textbf{
with the }\textbf{\emph{relative}}\textbf{ intra-class divergence
between benign images and their adversarial examples (}the adversarial
\emph{R-DIV} \textbf{$d_{r}^{+,adv}$)}. As shown in Figure \ref{fig:diver-acc-progress},
during the training process, the robust accuracy of the AT model tends
to improve, which corresponds with a decrease of the adversarial \emph{R-DIV}\textbf{
$d_{r}^{+,adv}$}. Similarly, when the robust accuracy of the NAT model
starts increasing at the epoch 100, the adversarial \emph{R-DIV}\textbf{
$d_{r}^{+,adv}$} starts decreasing. In addition, the
robust accuracy of the AT model is significantly higher than that of the NAT
model, whilst its \textbf{$d_{r}^{+,adv}$ }is far lower than that
of the NAT model. In Figure \ref{fig:diver-acc}, we visualize the
correlation between the R-DIV and the robust accuracy by generating
different attack strengths. It can be seen that there is a common
trend such that the lower robust accuracy the higher R-DIV, regardless
of the model architecture or defense methods. These observations support
our claim of the relation between robust accuracy and R-DIV.

(O2) In Figure \ref{fig:diver-compare-absolute}, we visualize the
absolute intra-class divergence ($d_{a}^{+}$) and the absolute inter-class
divergence ($d_{a}^{-}$) for the cases of the NAT/AT models with
their corresponding robust accuracies. It can be observed that: (i)
in the same architecture, the $d_{a}^{+}$ of the AT model is much smaller
than that of NAT model. However, the $d_{a}^{-}$ of the AT model is also
much smaller than that of NAT model. It implies that, the AT method
helps to compact the representations of intra-class samples, but undesirably
makes the representations of inter-class samples closer. (ii)
Overall, the relative intra-class divergence of the AT model is smaller
than that of the NAT model -- which might explain why the NAT model is easy to
be attacked, and again confirms our O1.

\paragraph{Conclusions from the observations. }

\cite{mao2019metric} and \cite{bui2020improving}
reached a conclusion that the\emph{ absolute adversarial intra-class
divergence} $d_{a}^{+}$ is a key factor for robustness against adversarial
examples. However, as indicated by our O1, it is only one side of
the coin. The reason is that the \emph{absolute adversarial intra-class
divergence }only cares about how far adversarial examples of a class
are from their counterpart benign images, and does not pay attention to the
inter-divergence to other classes. As analysed in our observation of O2-i, low $d_{a}^{-}$
possibly harms the robust accuracy, because in this case, adversarial examples of
other classes are very close to those of the given class.
This further indicates that the
\emph{absolute adversarial inter-class divergence} $d_{a}^{-}$ needs
to be taken into account and it is necessary to minimize the \emph{relative adversarial
intra-class divergence} $d_{r}^{+}=\frac{d_{a}^{+}}{d_{a}^{-}}$ better
controls both \emph{the absolute adversarial intra-class divergence}
and \emph{absolute adversarial inter-class divergence }for strengthening
robustness.
The above analytical and empirical study confirms the feasibility of applying SCL to enhance robustness in AML but one can also see that it is non-trivial to develop an appropriate strategy to be the combination effective.

\section{Proposed method}

In this section, we provide the answer for the question ``\textbf{\emph{How
}}\emph{to integrate CL with adversarial training in the context of
AML?}''.   We first propose an adapted version of SCL which we
call Adversarial Supervised Contrastive Learning (ASCL) for the AML
problem. We then introduce three sample selection strategies to nominate
the most relevant positives and negatives to the anchor, which further
improve robustness with much fewer samples.  

\subsection{Adversarial Supervised Contrastive Learning\label{subsec:ASCL}}

\paragraph{Terminologies.}

We consider a prediction model $h(\bx)=g(f(\bx))$ where $f()$ is
the encoder which outputs the latent representation $\bz=f(\bx)$
and $g()$ is the classifier upon the latent $\bz$. Also we have
a batch of N pairs $\{\bx_{i},\by_{i}\}_{i=1}^{N}$ of benign images
and their labels. With an adversarial transformation $\mathcal{A}$
(e.g., PGD),
each pair $\{\bx_{i},\by_{i}\}$ has two corresponding sets, a positive
set $\bX_{i}^{+}=\{\bx_{j},\bx_{j}^{a}\mid j\neq i,\by_{j}=\by_{i}\}$
and a negative set $\bX_{i}^{-}=\{\bx_{j},\bx_{j}^{a}\mid j\neq i,\by_{j}\neq\by_{i}\}$.
We then have the corresponding sets in the latent space $\bZ_{i}^{+}=\{f(\bx_{j})\mid\bx_{j}\in\bX_{i}^{+}\}$
and $\bZ_{i}^{-}=\{f(\bx_{j})\mid\bx_{j}\in\bX_{i}^{-}\}$.

\paragraph{Supervised Contrastive Loss.}

The supervised contrastive loss for an anchor $\bx_{i}$ as follow:
\begin{equation}
\mathcal{L}_{i}^{\text{scl}}=\frac{-1}{\left|\bZ_{i}^{+}\right|+1}\underset{\bz_{j}\in\bZ_{i}^{+}\cup\{\bz_{i}^{a}\}}{\sum}\log\frac{e^{\frac{sim(\bz_{j},\bz_{i})}{\tau}}}{\underset{\bz_{k}\in\bZ_{i}^{+}\cup\bZ_{i}^{-}\cup\{\bz_{i}^{a}\}}{\sum}e^{\frac{sim(\bz_{k},\bz_{i})}{\tau}}}\label{eq:scl-nat}
\end{equation}

where  $sim(\bz_{j},\bz_{i})$ represents the similarity metric between
two latent representations and $\tau$ is a temperature parameter.
It is worth noting that there are two changes in our SCL loss compared
with the original one in \cite{khosla2020supervised}. Firstly, $sim(\bz_{j},\bz_{i})$
is a general form of similarity, which can be any similarity metric
such as cosine similarity $\frac{\bz_{j}\cdot\bz i}{\left\Vert \bz_{j}\right\Vert \times\left\Vert \bz_{i}\right\Vert }$
or Lp norm $-\left|\bz_{j}-\bz_{i}\right|_{p}$. Secondly, in term
of terminology, in \cite{khosla2020supervised}, the positive set
was defined including those samples in the same class with the anchor
$\bx_{i}$ (e.g. $\bX_{i}^{+}$) and the anchor's transformation $\bx_{i}^{a}$.
However, in our paper, we want to emphasize the importance of the
anchor's transformation, therefore, we use two separate terminologies
$\bX_{i}^{+}$ and $\{\bx_{i}^{a}\}$. Similarly, the SCL loss for
an anchor $\bx_{i}^{a}$ as follow: 

\begin{equation}
\mathcal{L}_{i}^{\text{a,scl}}=\frac{-1}{\left|\bZ_{i}^{+}\right|+1}\underset{\bz_{j}\in\bZ_{i}^{+}\cup\{\bz_{i}\}}{\sum}\log\frac{e^{\frac{sim(\bz_{j},\bz_{i})}{\tau}}}{\underset{\bz_{k}\in\bZ_{i}^{+}\cup\bZ_{i}^{-}\cup\{\bz_{i}\}}{\sum}e^{\frac{sim(\bz_{k},\bz_{i})}{\tau}}}\label{eq:scl-adv}
\end{equation}

The average SCL loss over a batch is as follows: 

\begin{equation}
\mathcal{L}^{\text{SCL}}=\frac{1}{N}\sum_{i=1}^{N}\left(\mathcal{L}_{i}^{\text{scl}}+\mathcal{L}_{i}^{\text{a,scl}}\right)
\end{equation}

As mentioned in \cite{khosla2020supervised}, there is a major advantage
of SCL compared with Self-Supervised CL (SSCL) in the context of regular
machine learning. Unlike SSCL in which each anchor has only single
positive sample, SCL takes advantages of the labels to have many positives
in the same batch size N. This strategy helps to reduce the false
negative cases in SSCL when two samples in the same class are pushed
apart. As shown in \cite{khosla2020supervised}, SCL training
is more stable than SSCL and also achieves better performance. 

\paragraph{Adaptations in the context of AML.}

However, original SCL is not sufficient to achieve adversarial robustness.
In the context of adversarial machine learning, we need the following
adaptations to improve the adversarial robustness: 

(i)  As shown in Table 1 in \cite{kim2020adversarial}, 
the original SCL slightly improves the robustness of a standard model but 
cannot defend strong adversarial attacks. Therefore, we use an adversary $\mathcal{A}$
(e.g., PGD) as the transformation $\mathcal{T}$ instead
of the traditional data augmentation (e.g., combination of random
cropping and random jittering) as in other contrastive learning frameworks
\citep{chen2020simple,khosla2020supervised,he2020momentum}. This helps
to reduce the divergence in latent representations of a benign image
and its adversarial example directly. 

(ii) We apply SCL as a regularization on top of the Adversarial Training
(AT) method \citep{madry2017towards,Zhang2019theoretically,shafahi2019adversarial,xie2020smooth}.
Therefore, instead of pre-training the encoder $f()$ with contrastive
learning loss as in previous work, we can optimize the AT and the
SCL simultaneously. The AT objective function with the cross-entropy
loss $\mathcal{C}()$ is as follows: 

\begin{equation}
\mathcal{L}^{AT}=\frac{1}{N}\sum_{i=1}^{N}\mathcal{C}\left(h(\bx_{i}),\by_{i}\right)+\mathcal{C}\left(h(\bx_{i}^{a}),\by_{i}\right)
\end{equation}

\paragraph{Regularization on the prediction space.}

The clustering assumption \citep{chapelle2005semi} is a technique that
encourages the classifier to preserve its predictions for data examples
in a cluster. Theoretically, the clustering assumption enforces the
decision boundary of a given classifier to lie in the gap among the
data clusters and never cross over any clusters.  As shown in \cite{chen2020simple,khosla2020supervised},
with the help of CL,  latent representations of those samples in
the same class form into clusters. Therefore, coupling our SCL framework
with the clustering assumption can help to increase the margin from
a data sample to the decision boundary. 
To enforce the clustering assumption, we use Virtual Adversarial Training (VAT) \citep{VAT}
to maintain the classifier smoothness: 

\begin{equation}
\mathcal{L}^{\text{VAT}}=\frac{1}{N}\sum_{i=1}^{N}D_{KL}\left(h(\bx_{i})\parallel h(\bx_{i}^{a})\right)
\end{equation}

\paragraph{Putting it all together. }

We combine the relevant terms to the final objective function of our
framework which we name as Adversarial Supervised Contrastive Learning
(ASCL) as follows: 

\begin{equation}
\mathcal{L}=\mathcal{L}^{\text{AT}}+\lambda^{\text{SCL}}\mathcal{L}^{\text{SCL}}+\lambda^{\text{VAT}}\mathcal{L}^{\text{VAT}}
\end{equation}

where $\lambda^{\text{SCL}}$ and $\lambda^{\text{VAT}}$ are hyper-parameters
to control the SCL loss and VAT loss, respectively. As mentioned in
the observation (O2), minimizing the AT loss $\mathcal{L}^{AT}$ alone
compresses not only the representations of intra-class clusters but
also reduces the inter-class distance, which hurts the natural discrimination.
Therefore, coupling with $\mathcal{L}^{SCL}$ can compensate the aforementioned
weakness by simultaneously minimizing the intra-class divergence and
maximizing the inter-class divergence. Finally, by forcing predictions
of intra-class samples to be close, the VAT regularization $\mathcal{L}^{VAT}$
help to maintain the classifier smoothness and further improve the
robustness. In addition to the intuitive analysis, we also provide
an empirical ablation study to further understand the contribution
of each component in the supplementary material. 

\subsection{Global and Local Selection Strategies \label{subsec:Local-ASCL}}

\begin{table*}
    \caption{Definitions of positives and negatives with Global Selection and Local
    Selection strategies given an anchor $\{\protect\bx_{i},\protect\by_{i}\}$
    and a predicted label $p=\text{argmax}\;h(\protect\bx)$, $p^{a}=\text{argmax}\;h(\protect\bx^{a})$\label{tab:3locals}}
    \begin{centering}
    \resizebox{1.0\textwidth}{!}{\centering\setlength{\tabcolsep}{2pt}
    \begin{tabular}{c|c|c}
    \hline 
    & $\bX_{i}^{+}$ & $\bX_{i}^{-}$\tabularnewline
    \hline 
    Global & $\{\bx_{j},\bx_{j}^{a}\mid j\neq i,\by_{j}=\by_{i}\}$ & $\{\bx_{j},\bx_{j}^{a}\mid j\neq i,\by_{j}\neq\by_{i}\}$\tabularnewline
    Hard-LS & $\{\bx_{j},\bx_{j}^{a}\mid j\neq i,\by_{j}=\by_{i}\}$ & $\{\bx_{j}\mid j\neq i,\by_{j}\neq\by_{i},p_{j}=\by_{i}\}\cup\{\bx_{j}^{a}\mid j\neq i,\by_{j}\neq\by_{i},p_{j}^{a}=\by_{i}\}$\tabularnewline
    Soft-LS & $\{\bx_{j},\bx_{j}^{a}\mid j\neq i,\by_{j}=\by_{i}\}$ & $\{\bx_{j}\mid j\neq i,\by_{j}\neq\by_{i},p_{j}=p_{i}\}\cup\{\bx_{j}^{a}\mid j\neq i,\by_{j}\neq\by_{i},p_{j}^{a}=p_{i}\}$\tabularnewline
    Leaked-LS & $\{\bx_{j}\mid j\neq i,\by_{j}=\by_{i},p_{j}=p_{i}\}\cup\{\bx_{j}^{a}\mid j\neq i,\by_{j}=\by_{i},p_{j}^{a}=p_{i}\}$ & $\{\bx_{j}\mid j\neq i,\by_{j}\neq\by_{i},p_{j}=p_{i}\}\cup\{\bx_{j}^{a}\mid j\neq i,\by_{j}\neq\by_{i},p_{j}^{a}=p_{i}\}$\tabularnewline
    \hline 
    \end{tabular}}
    \par\end{centering}
    \end{table*}

\paragraph{Global Selection.}

The SCL as in Equations \ref{eq:scl-nat},\ref{eq:scl-adv} can be
understood as SCL with a Global Selection strategy, where each anchor
$\bx_{i}$ takes all other samples in the current batch into account
and splits them into a positive set $\bX_{i}^{+}$ and a negative
set $\bX_{i}^{-}$. For example, as illustrated in Figure \ref{fig:scl},
given an anchor, with the help of SCL, it will push away all negatives
and pull all positives regardless of their correlation in the space.
However, there are two issues of this strategy: 

(I1) The high inter-class divergence issue of a diverse dataset. Specifically,
there are true negative (but uncorrelated) samples which are very
different in appearance (e.g., an anchor-dog and negative samples-sharks)
and latent representations. Therefore, pushing them away does not
make any contribution to the learning other than making it more unstable.
The number of uncorrelated negatives is increased when the dataset
is more diverse. 

(I2) The high intra-class divergence issue when the dataset is very
diverse in some classes. For example, a class ``dog'' in the ImageNet
dataset may include many sub-classes (breeds) of dog. Specifically,
there are true positive (but uncorrelated) samples which are in the
same class with the anchor but different in appearance. In the context
of AML, two samples in the same class (e.g., ``dog'') can be attacked
to be very different classes (e.g., one to the class ``cat'', one
to the class ``shark''), therefore the latent representations of
their adversarial examples are even more uncorrelated. 

\paragraph{Local Selection.}
Based on the above analysis, we leverage label supervision to
propose a series of Local Selection (LS) strategies for the SCL framework,
which consider \emph{local and important }samples only and ignore
other samples in the batch as illustrated in Figure \ref{fig:scl}.
They are \emph{Hard-LS, Soft-LS }and \emph{Leaked-LS} as defined in
Table \ref{tab:3locals}. 

More specifically, in Hard-LS and Soft-LS, we consider the same set
of positives as in Global Selection. However, we filter out the true
negative but uncorrelated samples by only considering those are predicted
as similar to the anchor's true label (Hard-LS) or to the anchor's
predicted label (Soft-LS). These two strategies are to deal with the
issue (I1) by choosing negative samples that have most correlation
with the current anchor. Because they are very close in prediction
space, their representation is likely high correlated with the anchor's
representation. 

In Leaked-LS, we add an additional constraint on the positive set
to deal with the issue (I2). Specifically, we filter out the true
positive but uncorrelated samples by only choosing those are currently
predicted as similar to the anchor's prediction. It is worth noting
that, the additional constraint is applied on the positive set $\bX_{i}^{+}$
only. It means that, each anchor $\bx_{i}$ and its adversarial example
$\bx_{i}^{a}$ are always pulled close together. However, instead
of pulling all other positive samples in current batch, we only pull
those samples which are close with the anchor's representation to further
support and stabilize the contrastive learning. 

From a practical perspective, as later shown in the experimental section,
ASCL with Leaked-Local Selection (Leaked ASCL) improves the robustness
over that with Global Selection most notably, and with much fewer
positive and negative samples. It has been shown that, optimal negative 
samples for contrastive learning are task-dependent which guide 
representations towards task-relevant features that improve 
performance \citep{Tian0PKSI20, frankle2020all}. However, while these 
previous works focused on unsupervised-setting, our Local-ASCL is the 
first work to leverage supervision to select not only optimal negative 
samples but also optimal positive samples for robust classification task.

\section{Experiments}

In this section, we first introduce the experimental setting for adversarial attacks and defenses. 
We then provide an extensive robustness evaluation between our best method (which is Leaked-ASCL) 
with other defenses to demonstrate the significant improvement of ours. 
Finally, we empirically answer the question ``\textbf{\emph{What}}\emph{ are the important factors 
for the application of the ASCL framework in the context of AML?}'' through our experiments. 
We provide a comparison among Global/Local Selection strategies and show that the Leaked-ASCL not only 
outperforms the Global ASCL but also makes use of much fewer positives and negatives. 
An ablation study to investigate the importance of each component to the performance can be found in 
the supplementary material.

\subsection{Experimental Setting\label{subsec:Experimental-Setting}}

\paragraph{General Setting.}

We use CIFAR10 and CIFAR100 \citep{krizhevsky2009learning} as the benchmark datasets in our experiment. 
Both datasets have 50,000 training images and 10,000 test images. However, while the CIFAR10 dataset
has 10 classes, CIFAR100 is more diverse with 100 classes. 
The inputs were normalized to $[0,1]$. We apply random horizontal flips and
random shifts with scale $10\%$ for data augmentation as used in
\cite{pang2019improving}. 
We use four architectures including standard CNN, ResNet18/20 \citep{he2016deep} 
and WideResNet-34-10 \citep{zagoruyko2016wide} in our experiment. 
The architecture and training setting for each dataset are provided in
our supplementary material. 

\paragraph{Contrastive Learning Setting.}

We choose the penultimate layer ($l_{y}^{-1})$ as the intermediate
layer to apply our regularization. The analytical study for the effect
of choosing projection head in the context of AML can be found in
the supplementary material. In the main paper, we report the experimental
results without the projection head. The temperature $\tau=0.07$
as in \cite{khosla2020supervised}. 

\paragraph{Attack Setting.}

We use different state-of-the-art attacks to evaluate the defense
methods including: 
(i) \textbf{PGD attack }which is a gradient based attack. We use
$k=250,\epsilon=8/255,\eta=2/255$ for the CIFAR10 dataset and $k=250,\epsilon=0.01,\eta=0.001$
for the CIFAR100 dataset. We use two versions of the PGD attack: the
non-targeted PGD attack (PGD) and the multi-targeted PGD attack (mPGD). 
(ii) \textbf{Auto-Attack} \citep{croce2020reliable} which is an ensemble
based attack. We use $\epsilon=8/255$ for the CIFAR10 dataset and
$\epsilon=0.01$ for the CIFAR100 dataset, both with the standard
version of Auto-Attack (AA), which is an ensemble of four different
attacks.
The distortion metric we use in our experiments is $l_{\infty}$ for
all measures. We use the full test set for the attacks (i) and 1000
test samples for the attacks (ii).

\paragraph{Generating Adversarial Examples for Defenders.}

We employ PGD as the stochastic adversary to generate
adversarial examples. These adversarial examples have been used as
transformations of benign images in our contrastive framework. 
Specifically, the configuration for the CIFAR10 dataset
is $k=10,\epsilon=8/255,\eta=2/255$ and that for the CIFAR100 dataset
is $k=10,\epsilon=0.01,\eta=0.001$.

\paragraph{Baseline methods.}

Most closely related to our work is ADR \citep{bui2020improving}
which also aims to realize compactness in the latent space to improve
robustness in the supervised setting. We also compare with RoCL-TRADES \citep{kim2020adversarial} 
and ACL-DS \citep{jiang2020contrastive} which adversarially pre-trains with adversarial examples 
founded by Self-Supervised Contrastive loss and post-trains with standard supervised adversarial
training\footnote{The best reported version RoCL-AT-SS is a fine-tuned on 
a self-supervised ImageNet pretrained model, therefore, might not as a reference 
for comparison.}. 

\subsection{Robustness evaluation \label{subsec:Robustness-evaluation}}

\begin{table}
    \caption{Robustness evaluation on the CIFAR10 and CIFAR100 datasets with ResNet20
    architecture. Ours is Leaked-ASCL variant. GAP represents the average gap of robust accuracies
    between ours and the compared method.}
    \begin{centering}
    \resizebox{0.5\textwidth}{!}{
    \centering\setlength{\tabcolsep}{2pt}
        \begin{tabular}{cccccccccccc}
        \hline 
        & \multicolumn{5}{c}{CIFAR10} &  & \multicolumn{5}{c}{CIFAR100}\tabularnewline
        \cline{2-6} \cline{8-12}
        & Nat. & PGD & mPGD & AA & GAP &  & Nat. & PGD & mPGD & AA & GAP\tabularnewline
        \hline 
        ADV & \textbf{78.8} & 48.1 & 36.4 & 36.1 & 5.37 &  & \textbf{60.7} & 35.7 & 25.3 & 25.7 & 6.27\tabularnewline
        TRADES & \textbf{76.1} & 51.9 & 38.2 & 36.3 & 3.43 &  & \textbf{59.0} & 37.2 & 25.3 & 25.7 & 5.77\tabularnewline
        ADR & 76.8 & 51.5 & 38.9 & 38.6 & 2.57 &  & 59.1 & 40.0 & 29.1 & 28.6 & 2.60 \tabularnewline
        Ours & 75.5 & \textbf{53.7} & \textbf{41.0} & \textbf{42.0} & 0 &  & 59.0 & \textbf{42.5} & \textbf{31.1} & \textbf{31.9} & 0 \tabularnewline
        \hline 
        \end{tabular}}
    \par\end{centering}
    \label{tab:eval-cf10-cf100}
    \end{table}

\begin{table}
        \caption{Robustness evaluation against Auto-Attack with ResNet18 and
        WideResNet on the full test set of CIFAR10 dataset. 
        $\star$ Results are copied from \citet{croce2020robustbench}. 
        $\ddagger$ Results are copied from original papers, using a larger batch size (bs).  
        $\ast$ Omit the cross-entropy loss of natural images and VAT loss. Detail can be found
        in the supplementary material.}
        \centering \resizebox{0.5\textwidth}{!}{
            \begin{tabular}{llccc}
                \hline 
                    & Model & Nat & AA & PGD\tabularnewline
                \hline 
                Ours $\ast$ & WideResNet & 87.70 & 52.80 & 54.05\tabularnewline
                \citet{zhang2020attacks} $\star$& WideResNet & 84.52 & 53.51 & -\tabularnewline
                \citet{huang2020self} $\star$& WideResNet & 83.48 & 53.34 & -\tabularnewline
                \citet{Zhang2019theoretically} $\star$& WideResNet & 84.92 & 53.08 & -\tabularnewline
                \citet{cui2020learnable} $\star$& WideResNet & 88.22 & 52.86 & -\tabularnewline
                Ours $\ast$ & ResNet18 & 85.02 & 50.31 & 53.40\tabularnewline
                ACL-DS (bs=512) $\ddagger$& ResNet18 & 82.19 & - & 52.82\tabularnewline
                RoCL-TRADES (bs=256) $\ddagger$& ResNet18 & 84.55 & - & 43.85\tabularnewline
                \hline 
                \end{tabular}}     
        \label{tab:res-wrn}    
        \end{table}

\begin{figure}
    \begin{centering}
    \resizebox{0.7\linewidth}{!}{
    \includegraphics[width=0.8\columnwidth]{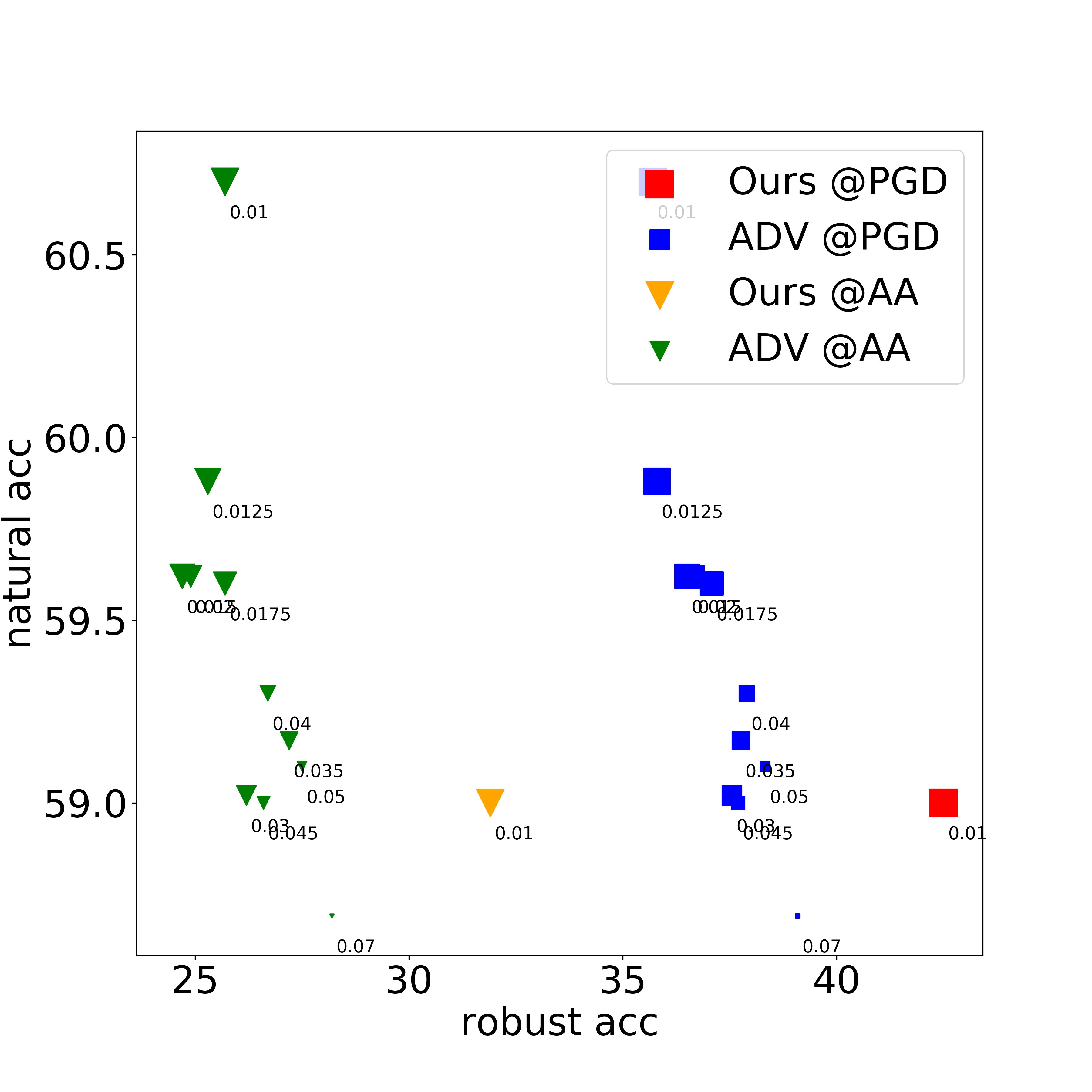}
    }
    \par\end{centering}
    \caption{Tradeoff between natural/robust accuracies when increasing perturbation
    magnitude (specified beside markers). For better visualization, bigger marker 
    indicates smaller perturbation. 
    Ours is Leaked-ASCL.}
    \label{fig:tradeoff}
    \vspace{-7mm}
    \end{figure}

We conduct extensive evaluations to demonstrate the advantages of our method (Leaked-ASCL variant) over 
other defenses. 
Table \ref{tab:eval-cf10-cf100} shows the robustness comparison 
on the CIFAR10 and CIFAR100 datasets with ResNet20 architecture. 
It can be seen that our method achieves much better robustness than the baseline methods on both datasets. 
More specifically, on the CIFAR10 dataset, the average gaps of robust accuracies against three attacks 
(PGD, mPGD and AA) between ours and ADR, TRADES, and ADV are $2.57\%$, $3.43\%$ and $5.37\%$, respectively. 
The similar gaps for the CIFAR100 dataset are $2.60\%$, $5.77\%$ and $6.27\%$, respectively. 
Figure \ref{fig:tradeoff} shows the tradeoff between natural accuracy and robust accuracies when increasing perturbation 
magnitude. It can be seen that simply increasing the magnitude of adversarial examples cannot
reach our performance even with a fine-range of perturbation. 
With the same level of natural accuracy, our method outperforms the baseline by around 
5\% which again emphasizes the advantage of our method. 
Finally, we compare our method with recently listed methods on the RobustBench \citep{croce2020robustbench} which have a similar setting 
(e.g., without additional data) as shown in Table \ref{tab:res-wrn}. With WideResNet architecture, our method achieves $52.80\%$ robust accuracy against Auto-Attack and 
87.70\% natural accuracy which is comparable with the SOTA method from \citet{cui2020learnable}. Compare to the best robust method 
from \citet{zhang2020attacks}, our method has 0.7\% lower in robust accuracy but 3.2\% higher in natural performance.   
With a smaller batch size, our method still achieves much better performance than RoCL and ACL 
which are two SOTA self-supervised contrastive learning based defenses.  

\subsection{Global and Local Selection strategies \label{subsec:exp-compare-global-local}}

In this subsection, we compare the effect of different global/local
selection strategies to the final performance. 
The comparison in Table \ref{tab:ab-local} shows that while the Hard-ASCL and Soft-ASCL
show a small improvement over ASCL, the Leaked-ASCL achieves the best
robustness compared with other strategies. 
\begin{figure}
    \begin{centering}
    \subfloat[Global/Local\label{fig:num-samples-all}]{\begin{centering}
    \includegraphics[width=0.5\columnwidth]{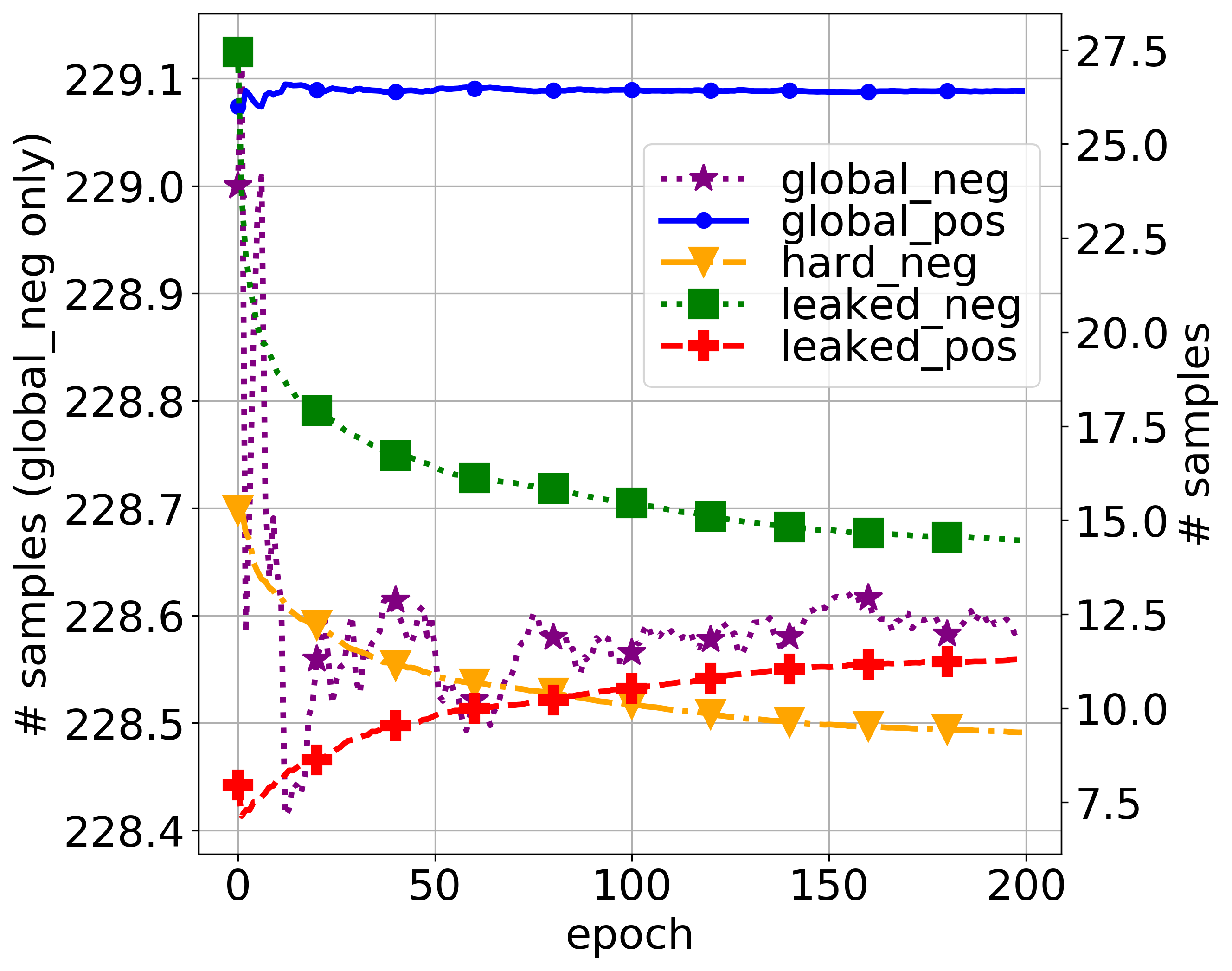}
    \par\end{centering}
    }\subfloat[Detail of Leaked Local\label{fig:num-samples-leaked}]{\begin{centering}
    \includegraphics[width=0.43\columnwidth]{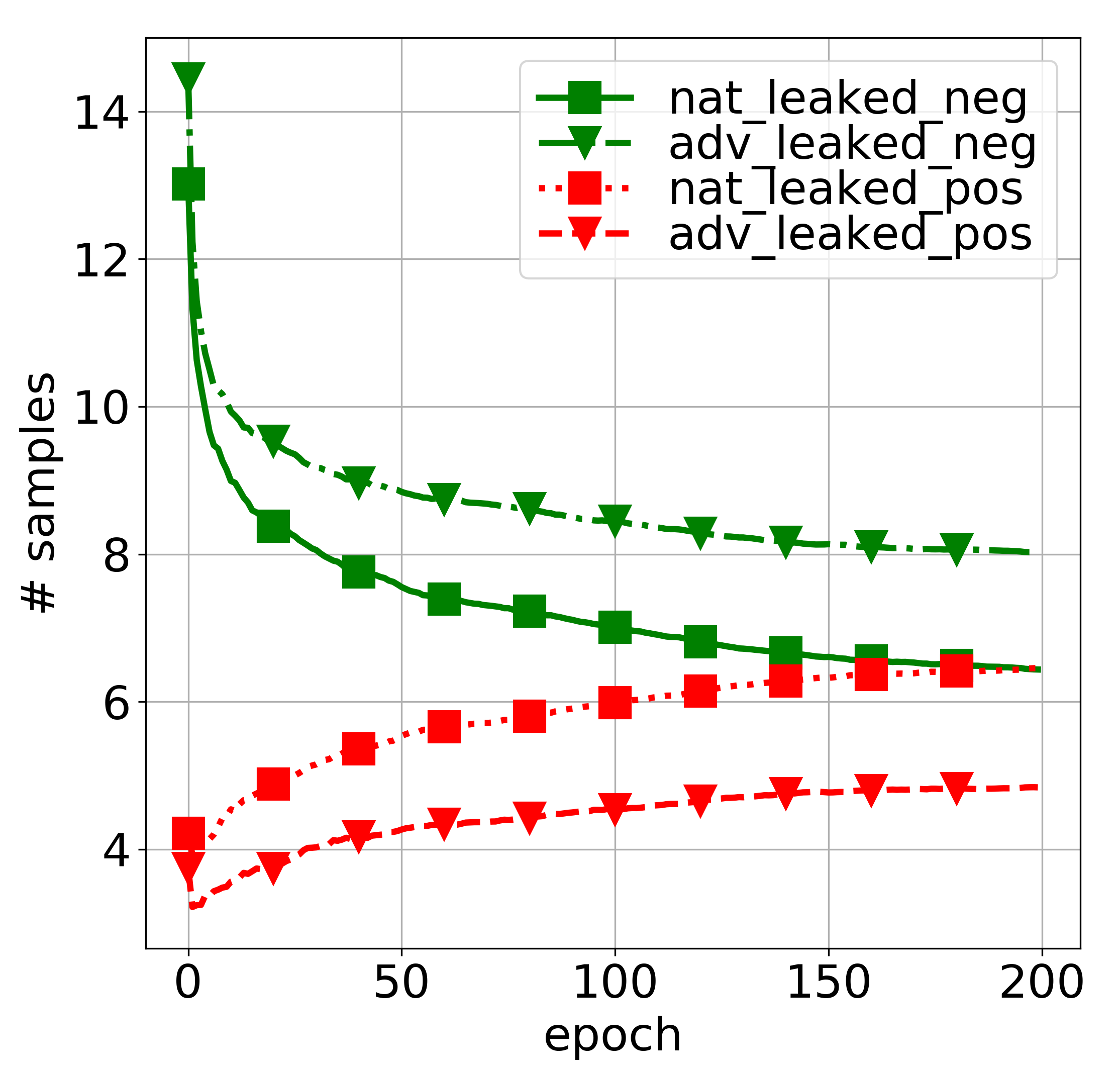}
    \par\end{centering}
    }
    \par\end{centering}
    \caption{Number of positives and negatives with different Global/Local Selection
    strategy on CIFAR10 dataset with batch size 128\label{fig:number-samples}}
\vspace{-5mm}    
\end{figure}
\begin{table}
    \caption{Comparison among Global/Local Selection Strategies on the CIFAR10
    dataset with ResNet20\label{tab:ab-local}}
    \begin{centering}
    \resizebox{0.3\textwidth}{!}{\centering\setlength{\tabcolsep}{2pt}
    \begin{tabular}{ccccc}
    \hline 
    & Nat. & PGD & mPGD & AA\tabularnewline
    \hline 
    (Global) ASCL & \textbf{76.4} & 52.7 & 40.4 & 40.9\tabularnewline
    Hard-ASCL & 75.5 & 53.1 & 41.0 & 41.3\tabularnewline
    Soft-ASCL & 75.5 & 53.4 & 40.6 & 40.4\tabularnewline
    Leaked-ASCL & 75.5 & \textbf{53.7} & \textbf{41.0} & \textbf{42.0}\tabularnewline
    \hline 
    \end{tabular}}
    \par\end{centering}
    \vspace{-7mm}
    \end{table}
We also measure the average number of positive and negatives samples
per batch corresponding with different selection strategies as shown
in Figure \ref{fig:num-samples-all}. With batch size 128, we have
a total of 256 samples per batch including benign images and their
adversarial examples. It can be seen that, the average positives and
negatives by the Global Selection are stable at 26.4 and 228.6, respectively.
In contrast, the number of positives and negatives by the Leaked-LS
vary corresponding with the current performance of the model. More
specifically, there are four advantages of the Leaked-LS over the
Global Selection: 

(i) at the beginning of training, approximately 7.5 positive samples
and 25 negative samples were selected. This is because of the low
classification performance of the model. Moreover, the strength of
the contrastive loss is directly proportional with the size of the
positive set. Therefore, with a small positive set, the contrastive
loss is weak in comparison with other components of ASCL. This helps
the model focuses more on improving the classification performance
first. 

(ii) when the model is improved, the number of positive samples is
increased, while the number of negative samples is decreased significantly. 
In addition to the bigger positive set, the contrastive loss
become stronger in comparison with other components. This helps the
model now focus more on contrastive learning and learning the
compact latent representation. 

(iii) unlike Global Selection, Leaked-LS considers
natural images and adversarial images differently based on their hardness
to the current anchor. As shown in Figure \ref{fig:num-samples-leaked},
there are more adversarial images than natural images in the negative
set, which helps the encoder focus to contrast the anchor with the
adversarial images. 

(iv) at the last epoch, Leaked-LS chooses only 11.3 positives and
14.3 negatives, which equate to $42.8\%$ and $6.3\%$ of the
positive set and negative set with the Global Selection strategy,
respectively. 
\vspace{-2mm}
\subsection{Why do ASCL and Local-ASCL improve adversarial robustness\label{subsec:exp-why-work}}

\begin{figure}
\begin{centering}
\subfloat[R-DIV\label{fig:mul-strengths-rdiv}]{\begin{centering}
\includegraphics[width=0.45\columnwidth]{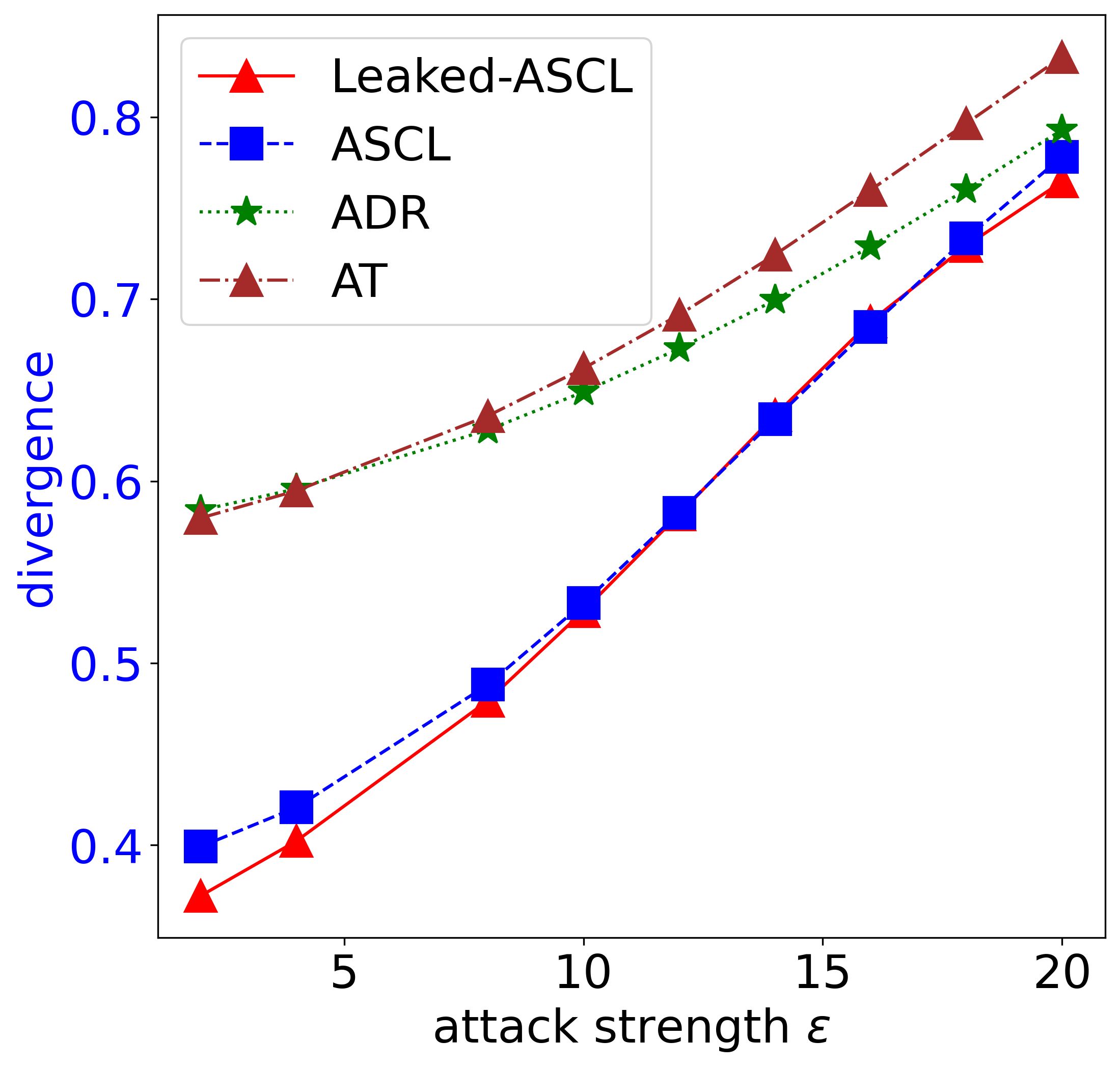}
\par\end{centering}
}\subfloat[Robust accuracy\label{fig:mul-strengths-acc}]{\begin{centering}
\includegraphics[width=0.45\columnwidth]{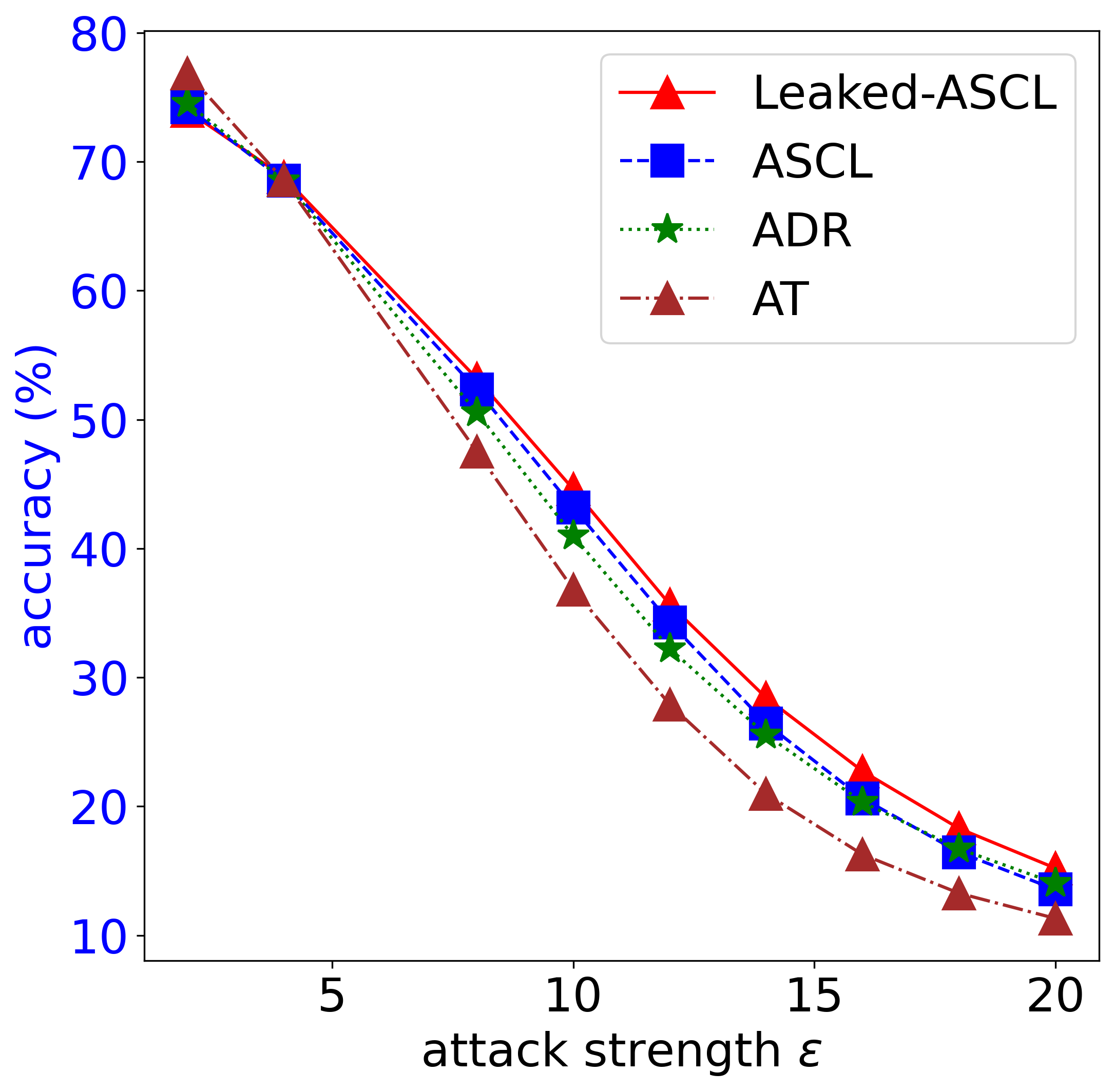}
\par\end{centering}
}
\par\end{centering}
\caption{R-DIV and robust accuracy under different attack strengths on CIFAR10
with ResNet20.\label{fig:R-DIV-and-robust}}
\vspace{-5mm}
\end{figure}

In this subsection, we connect with the hypothesis in Section \ref{sec:diver}
to explain why our ASCL and especially our Leaked-ASCL help to improve
adversarial robustness. Figure \ref{fig:R-DIV-and-robust} shows the
Relative intra-class divergence (R-DIV) and robust accuracy under
PGD attack $\{k=250,\eta=2/255\}$ with different attack strengths
$\epsilon$. It can be seen that (i) our ASCL and Leaked-ASCL have
lower R-DIV than baseline methods, and Leaked-ASCL achieves the lowest
measure, (ii) consequently, our ASCL and Leaked-ASCL achieves better
robust accuracy than baseline methods. Leaked-ASCL achieve the
best performance regardless of attack scenarios. The experimental
results concur with the proposed correlation between the Relative intra-class
divergence and the adversarial robustness as pointed out in Section
\ref{sec:diver}. Our methods help the representations of intra-class
samples to be more compact while increasing the margin between inter-class
clusters, and therefore improve the robustness.

\vspace{-2mm}
\section{Conclusion \label{sec:Conclusion}}

In this paper, we have shown the connection between robust accuracy
and the divergence in latent spaces. We demonstrated that 
contrastive learning 
can be applied to improve adversarial robustness
by reducing the intra-instance divergence while maintaining the inter-class
divergence. Moreover, we have shown that, instead of using all negatives
and positives as per the regular contrastive learning framework, by
judiciously picking highly correlated samples, we can further improve
the adversarial robustness.

\bibliography{ascl_21_ref}

\onecolumn
\clearpage{}
\begin{center}
\textbf{Supplementary material of ``Understanding and Achieving Efficient
Robustness with Adversarial Supervised Contrastive Learning''}
\par\end{center}

This supplementary material provides technical and experimental details as well as auxiliary aspects 
to complement the main paper. Briefly, it contains the following: 
\begin{itemize}
    \item Section \ref{sec:Training-setting}: Experimental setting.
    \item Section \ref{sec:add-latent-divergence}: Additional analysis of latent space divergence as well as 
    t-SNE visualization.
    \item Section \ref{sec:add-exp-results}: Additional experimental results which includes analysis of projection 
    head in Section \ref{sec:project-head}, contribution of each component in Section \ref{subsec:Ablation-study}, 
    image samples from Global and Local Selection strategies in Section \ref{subsec:add-Global-Local}. 
    \item  Section \ref{sec:related}: Discussion on the background, related works, and the necessity of our formulations for delivering our method.  
\end{itemize}
\section{Experimental setting \label{sec:Training-setting}}

\paragraph{General Setting.}

We use CIFAR10 and CIFAR100 \citep{krizhevsky2009learning} as the benchmark datasets in our experiment. 
Both datasets have 50,000 training images and 10,000 test images. However, while the CIFAR10 dataset
has 10 classes, CIFAR100 is more diverse with 100 classes. 
The training time is 200 epochs for both CIFAR10 and CIFAR100 datasets with batch size 128.
The inputs were normalized to $[0,1]$. We apply random horizontal flips and
random shifts with scale $10\%$ for data augmentation as used in \cite{pang2019improving}. 

We use four architectures including standard CNN, ResNet18/20 \citep{he2016deep} 
and WideResNet-34-10 \citep{zagoruyko2016wide} in our experiment. 
The standard CNN architecture has 4 convolution layers followed
by 3 FC layers as described in \cite{carlini2017towards}. 
For ResNet20 architectures, we use the same training setting as in \cite{pang2019improving}.
More specifically, we use Adam optimizer, with learning rate $10^{-3},10^{-4},10^{-5},10^{-6}$
at epoch 0th, 80th, 120th, and 160th, respectively. We use Adam optimization
with learning rate $10^{-3}$ for training the standard CNN architecture.

For ResNet18 and WideResNet architectures, we use the same training setting as in 
\cite{pang2020bag}. More specifically, we use SGD optimizer, with momentum $5 \times 10^{-4}$, 
with learning rate $10^{-1},10^{-2},10^{-3}$ at epoch 0th, 100th and 150th, respectively. 
It is a worth noting that, there are some modifications in the experiment in Table 3 to match the performance 
as in RobustBench: 
(i) we omit the cross-entropy loss of natural images in Eq. (4), so that 
the model sacrifices natural performance to gain more robust performance. The AT objective function 
becomes: $\mathcal{L}^{AT} = \frac{1}{N} \sum_{i=1}^{N} \mathcal{C} \left( h(\bx_i^a) , \by_i\right)$ which is 
similar as in \cite{pang2020bag}. 
(ii) we omit the VAT loss to show that the improvement truly comes from the contribution of the adversarial 
contrastive loss. 

\paragraph{Contrastive Learning Setting.}

We apply the contrastive learning on the intermediate layer $(l_{y}^{-1})$ which is 
intermediately followed by the last FC layer of either CNN or ResNet/WideResNet architectures.
The analytical study for the effect of choosing projection head in the context of AML can be 
found in Section \ref{sec:project-head}. 
In the main paper, we report the experimental results without the projection head. 
The temperature $\tau=0.07$ as in \cite{khosla2020supervised}. 

\paragraph{Attack Setting.}

We use different state-of-the-art attacks to evaluate the defense
methods including: 
(i) \textbf{PGD attack }which is a gradient based attack. We use
$k=250,\epsilon=8/255,\eta=2/255$ for the CIFAR10 dataset and $k=250,\epsilon=0.01,\eta=0.001$
for the CIFAR100 dataset. We use two versions of the PGD attack: the
non-targeted PGD attack (PGD) and the multi-targeted PGD attack (mPGD). 
(ii) \textbf{Auto-Attack} \citep{croce2020reliable} which is an ensemble
based attack. We use $\epsilon=8/255$ for the CIFAR10 dataset and
$\epsilon=0.01$ for the CIFAR100 dataset, both with the standard
version of Auto-Attack (AA), which is an ensemble of four different
attacks.
The distortion metric we use in our experiments is $l_{\infty}$ for
all measures. We use the full test set for the attacks (i) and 1000
test samples for the attacks (ii).

\paragraph{Generating Adversarial Examples for Defenders.}

We employ PGD as the stochastic adversary to generate
adversarial examples. These adversarial examples have been used as
transformations of benign images in our contrastive framework. 
Specifically, the configuration for the CIFAR10 dataset
is $k=10,\epsilon=8/255,\eta=2/255$ and that for the CIFAR100 dataset
is $k=10,\epsilon=0.01,\eta=0.001$.

\section{Additional Analysis of Latent Space Divergence \label{sec:add-latent-divergence}}

\paragraph{Experimental setting.}

The training setting has been described in Section \ref{sec:Training-setting}. 
Because the intra-class/inter-class divergences are averagely calculated
on all $N^{2}$ pairs of latent representations which is over our
computational capacity, therefore, we alternately calculate these
divergences on a mini-batch (128) and take the average over all mini-batches. 

\paragraph{Additional evaluation.}

\begin{figure}
\begin{centering}
\includegraphics[width=0.5\columnwidth]{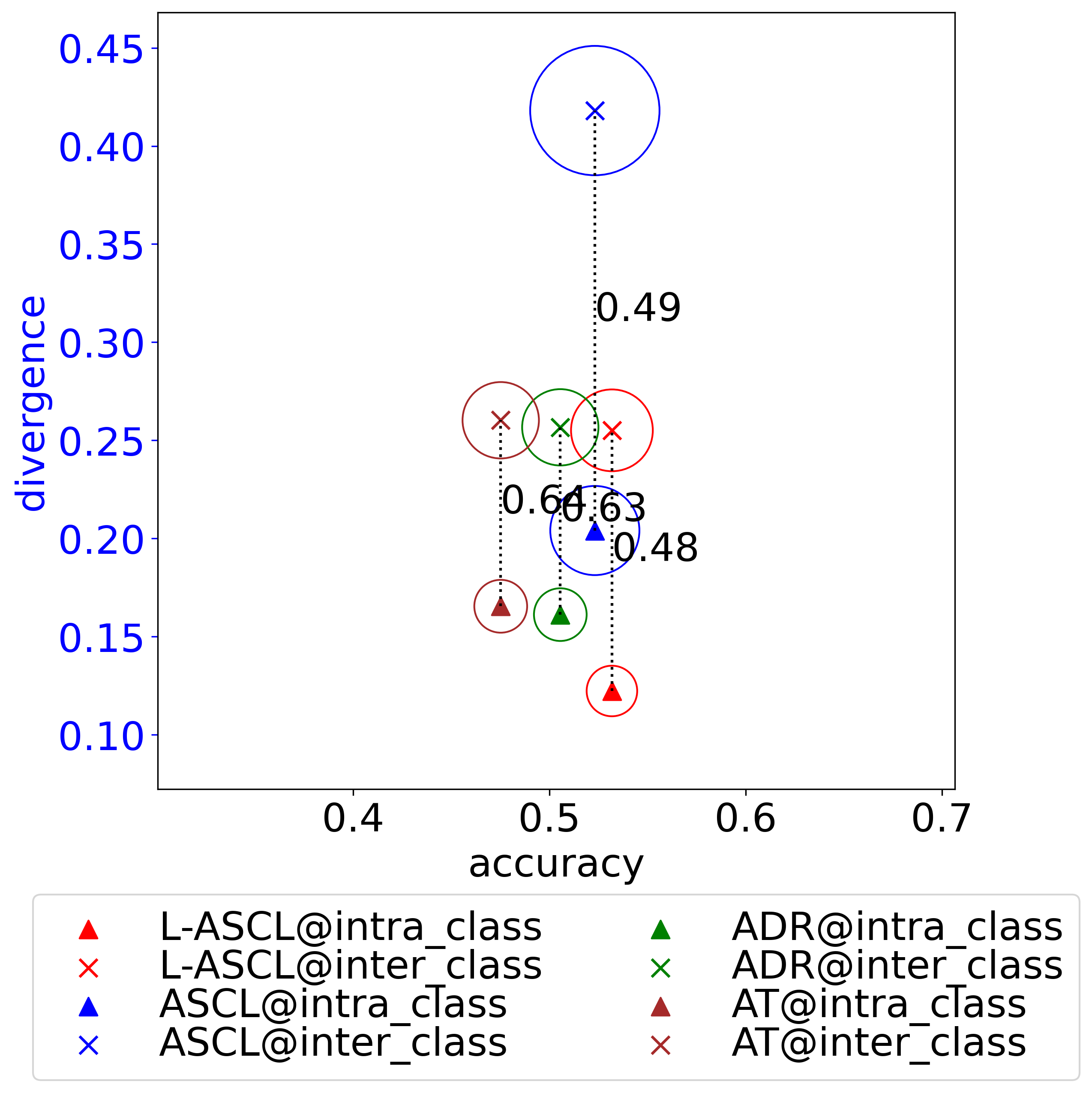}
\par\end{centering}
\caption{Pairs of Absolute-DIV with corresponding robust accuracy and R-DIV
(noted in each line).\label{fig:rdiv-acc-all}}
\end{figure}

In addition to the comparison in Section 4.2, we provide a further
evaluation on R-DIV and robust accuracy on the CIFAR10 dataset with
ResNet20 architecture, under PGD attack $\{\epsilon=8/255,\eta=2/255,k=250\}$
as shown in Figure \ref{fig:rdiv-acc-all}. It can be observed that
(i) the value of R-DIV decreases in order from AT (0.64), ADR (0.63),
ASCL (0.49), Leaked-ASCL (0.48), respectively. On the other hand,
the robust accuracy increases in the same order. (ii) ASCL has much
higher absolute intra-class divergence and inter-class divergence
than ADR and AT methods, however, ASCL has much lower R-DIV comparing
with two baseline methods, therefore, explaining its higher robust
accuracy. This result is similar with the comparison on Figure 2.a 
and the observation O2-i in the main paper and further confirm our
conclusion such that ``the robustness varies inversely with the relative
intra-class divergence between benign images and their adversarial
examples''.

\paragraph{t-SNE visualization.}

\begin{figure*}
\begin{centering}
\subfloat[NAT\label{fig:tsne-NAT}]{\begin{centering}
\includegraphics[width=0.35\columnwidth]{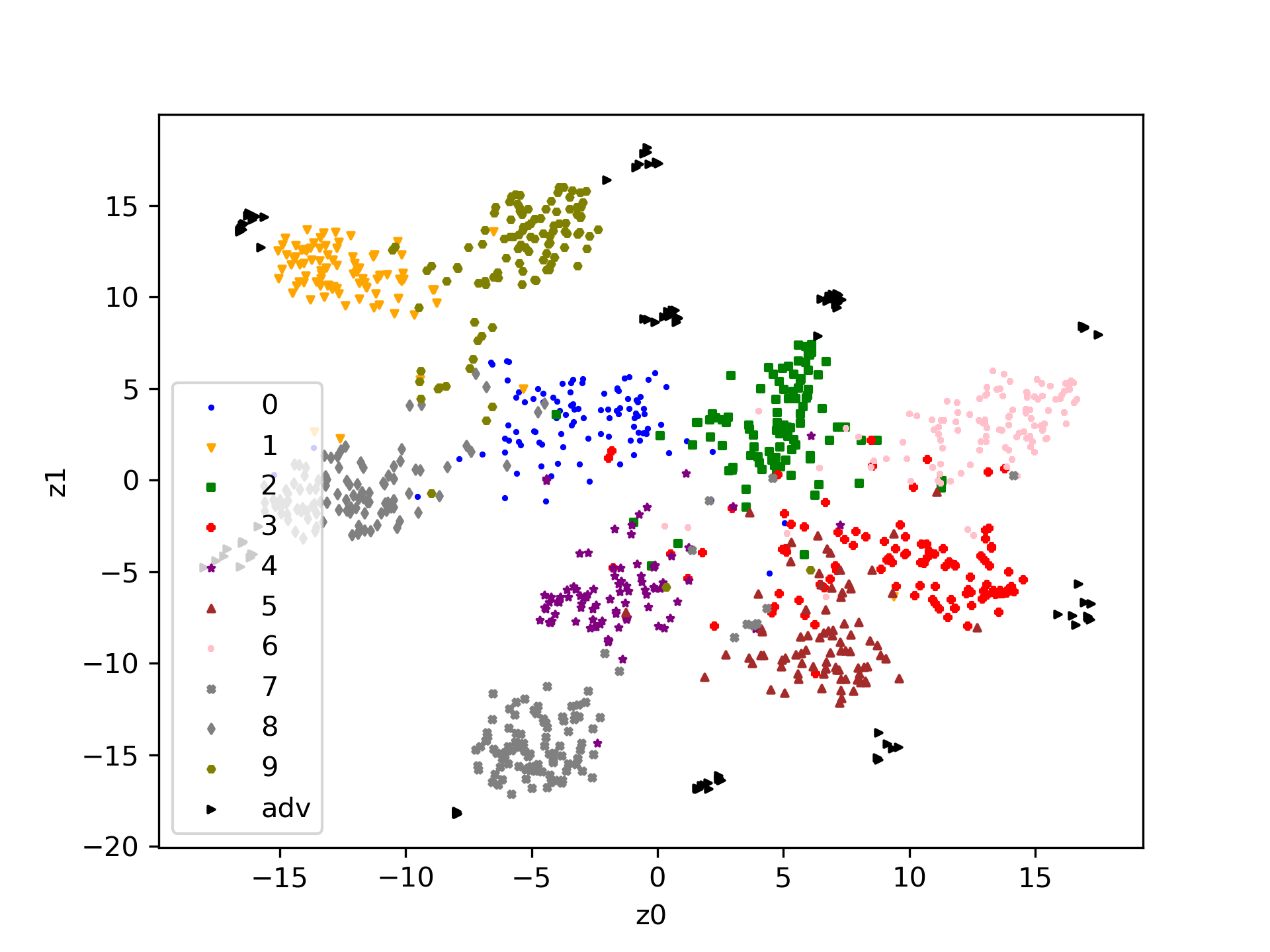}\includegraphics[width=0.35\columnwidth]{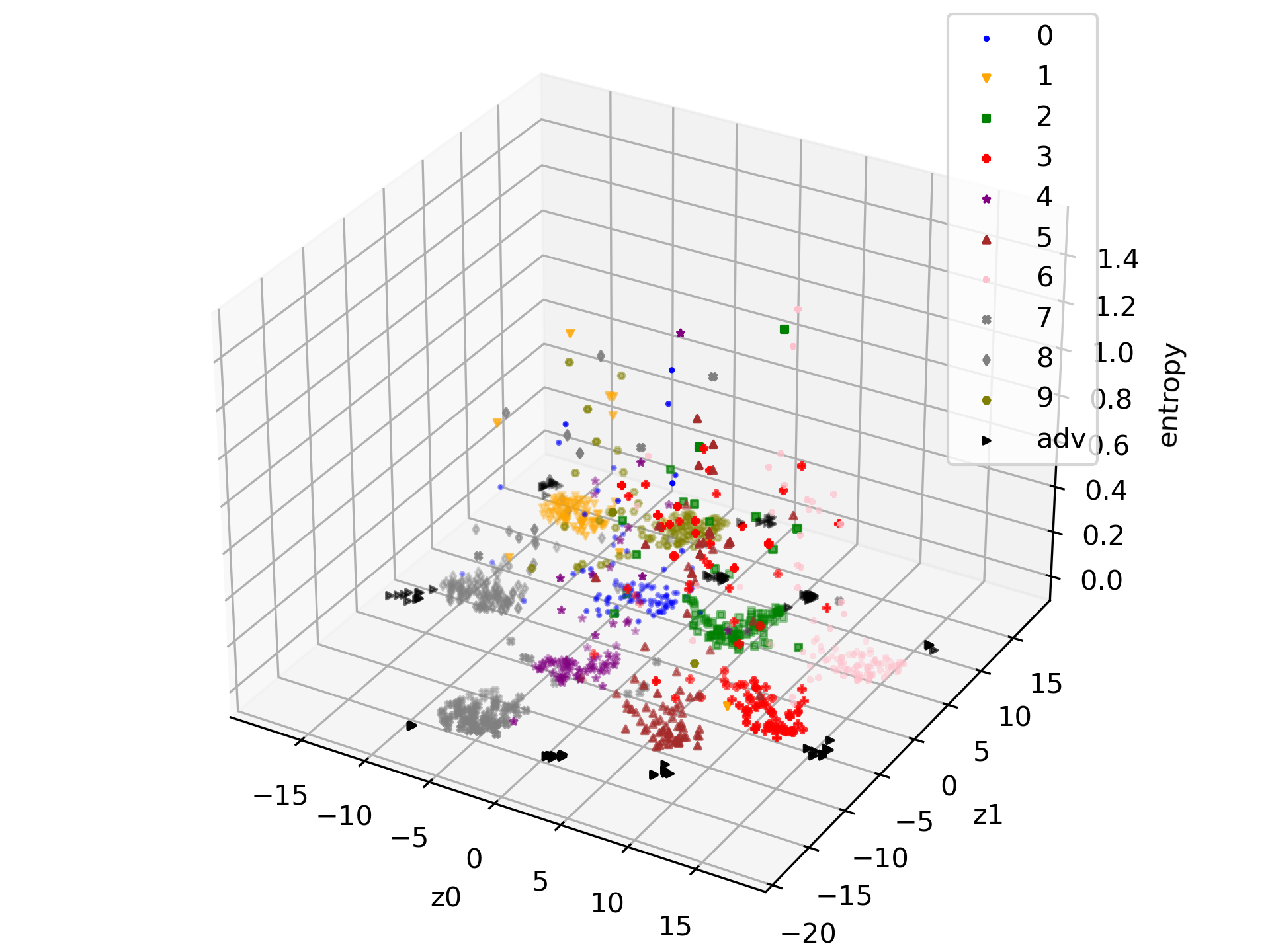}\vspace{-3mm}
\par\end{centering}
}
\par\end{centering}
\begin{centering}
\vspace{-3mm}
\subfloat[AT\label{fig:tnse-AT}]{\begin{centering}
\includegraphics[width=0.35\columnwidth]{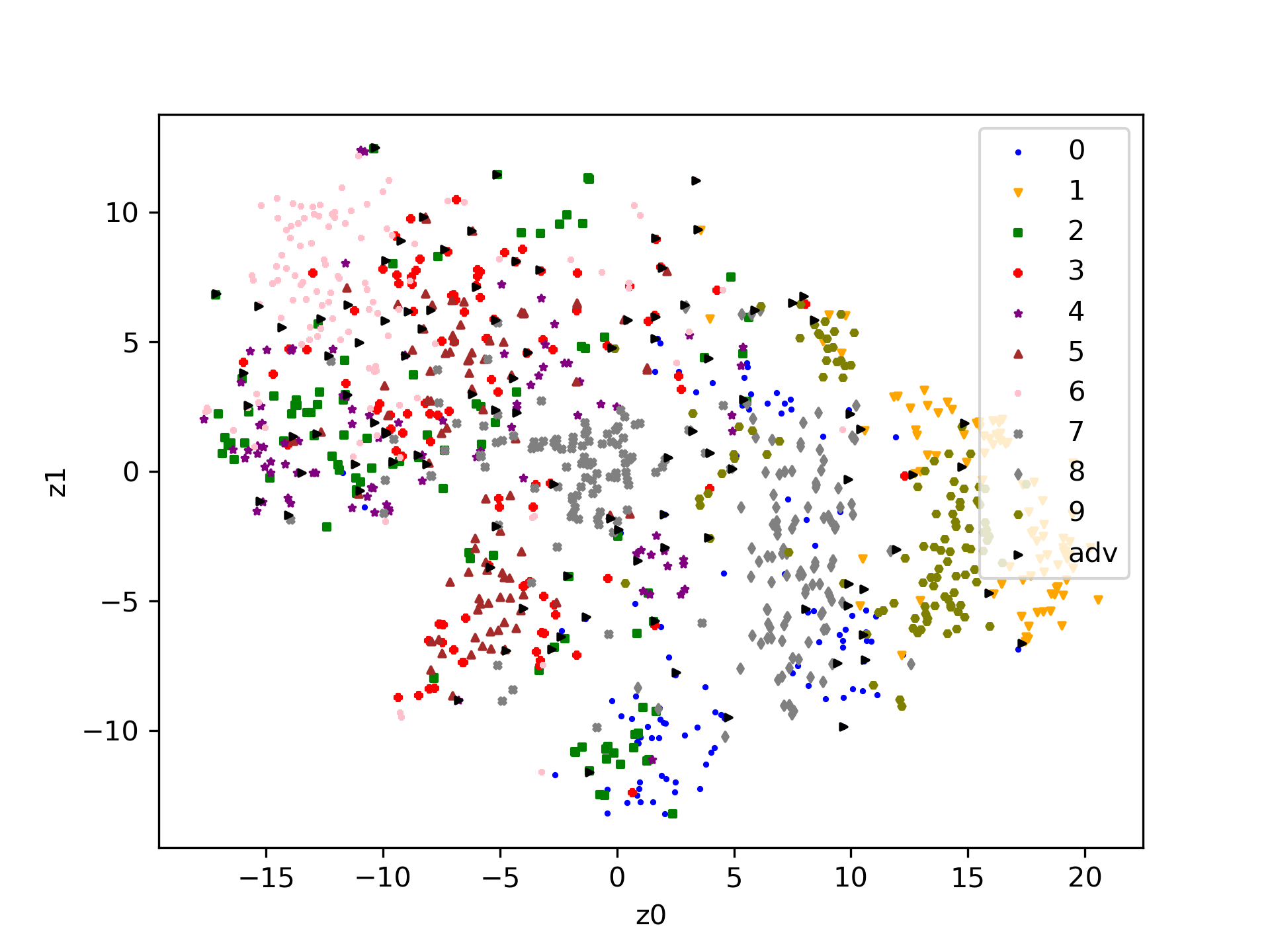}\includegraphics[width=0.35\columnwidth]{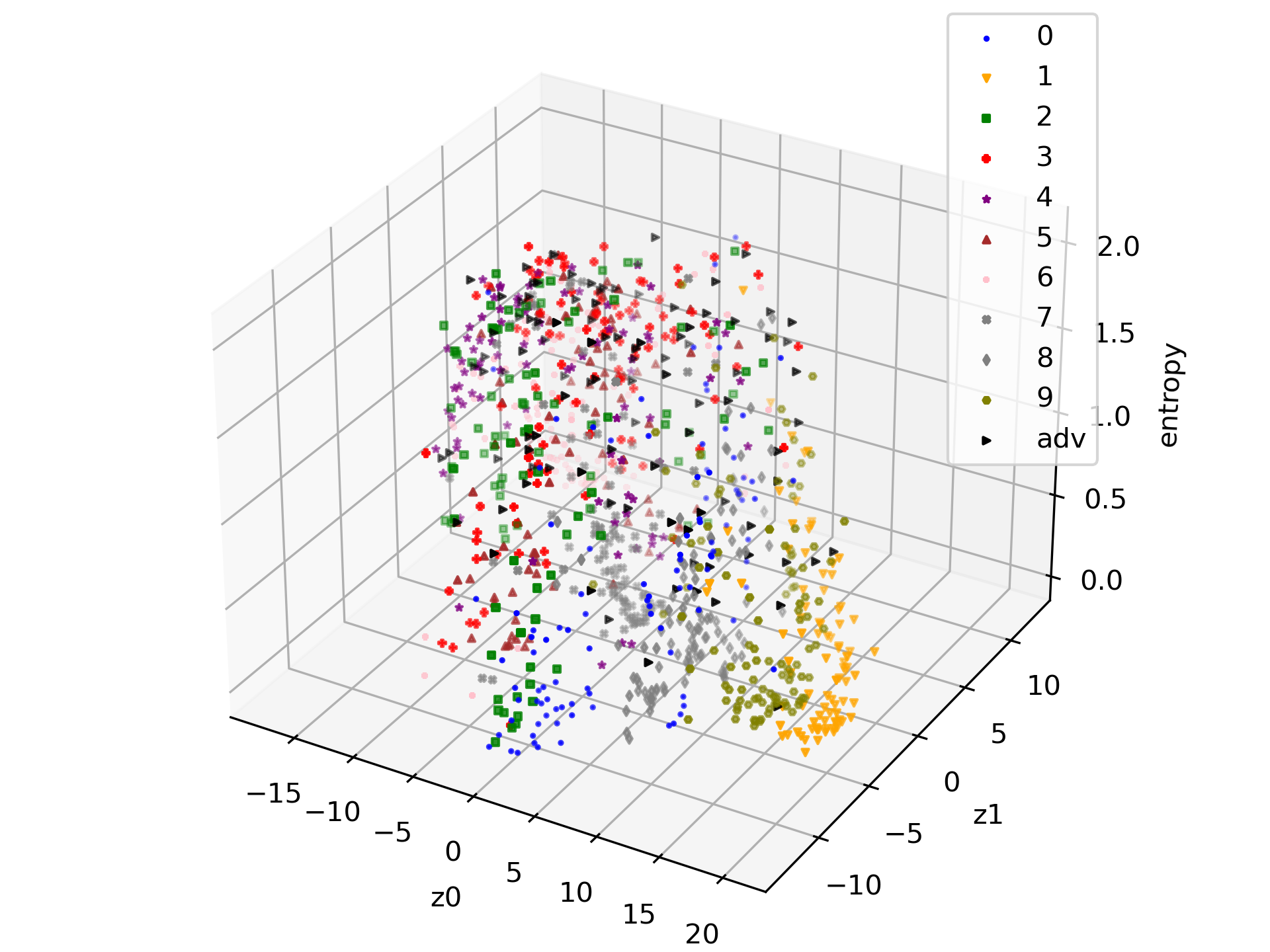}\vspace{-3mm}
\par\end{centering}
}
\par\end{centering}
\begin{centering}
\vspace{-3mm}
\subfloat[ASCL\label{fig:tnse-ASCL}]{\begin{centering}
\includegraphics[width=0.35\columnwidth]{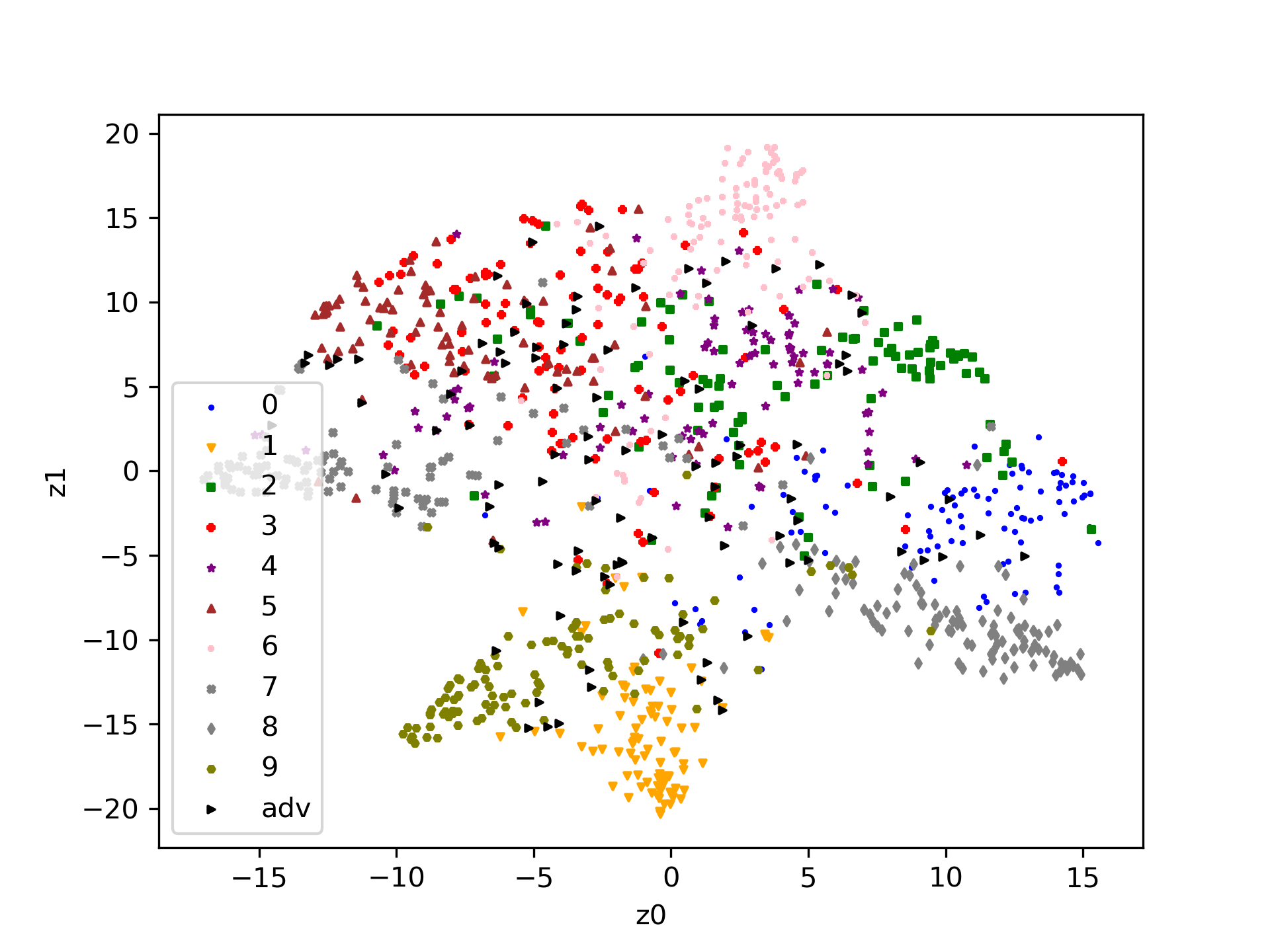}\includegraphics[width=0.35\columnwidth]{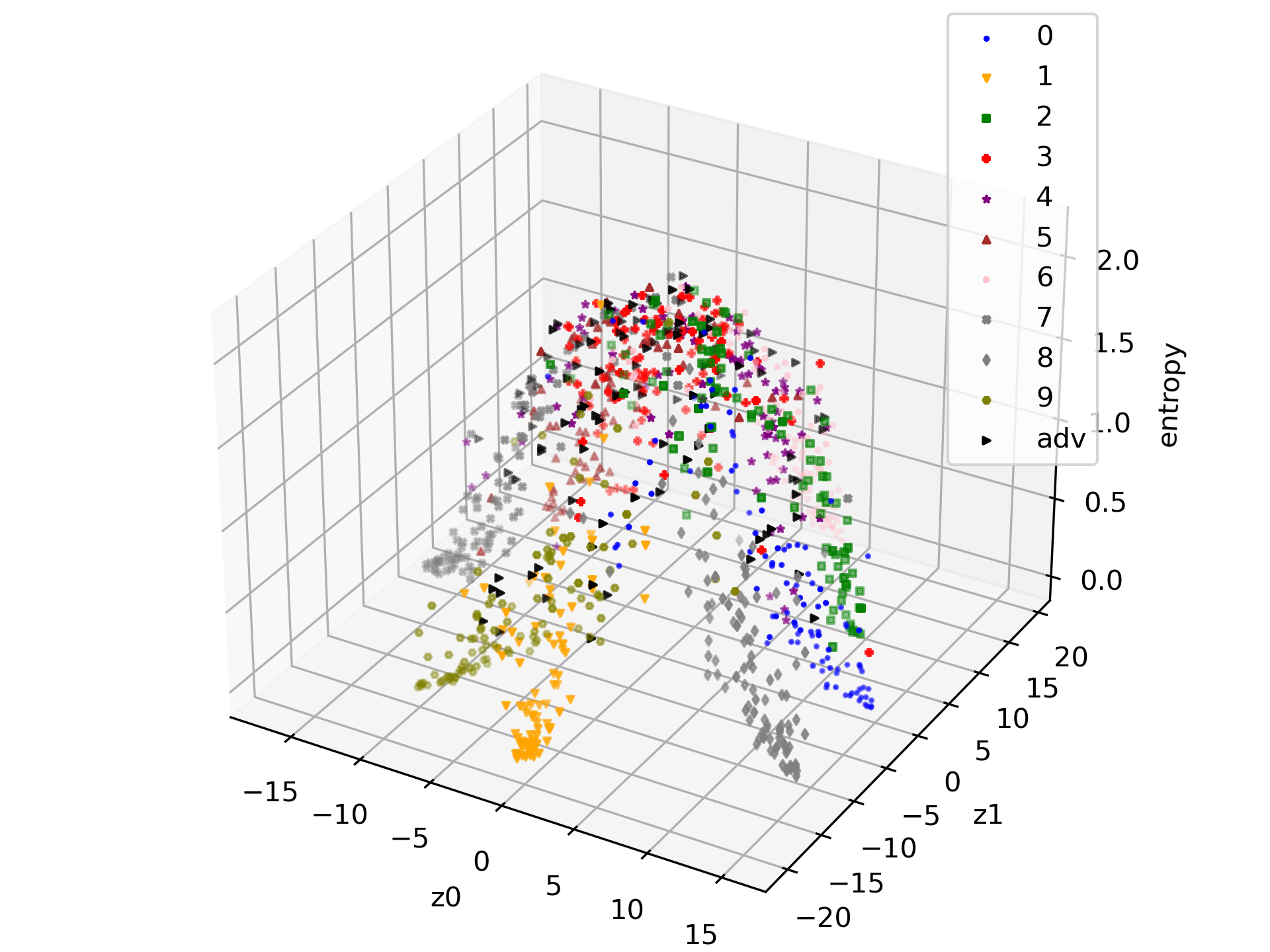}\vspace{-3mm}
\par\end{centering}
}\vspace{-3mm}
\par\end{centering}
\begin{centering}
\subfloat[Leaked-ASCL\label{fig:tnse-LASCL}]{\begin{centering}
\includegraphics[width=0.35\columnwidth]{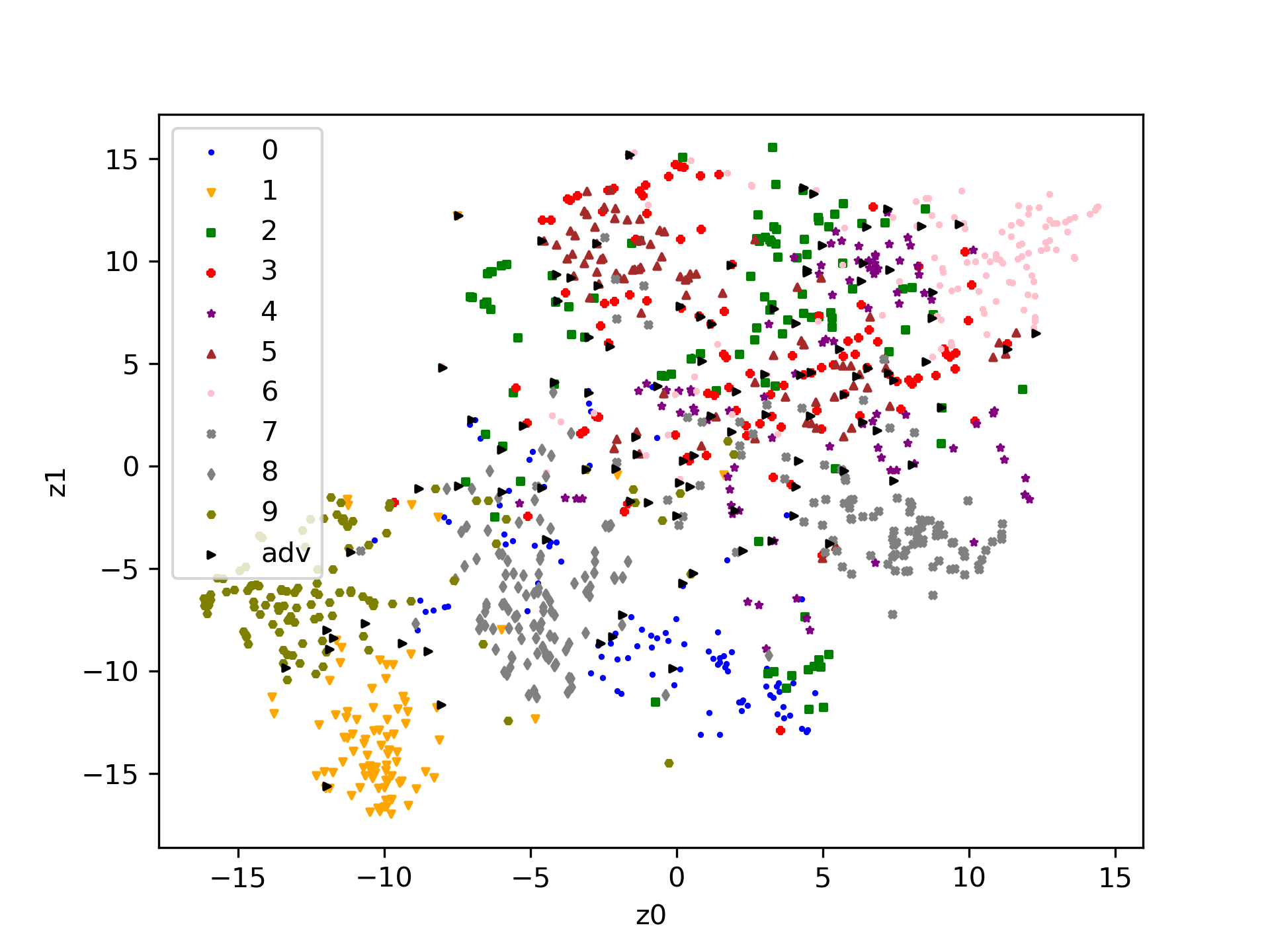}\includegraphics[width=0.35\columnwidth]{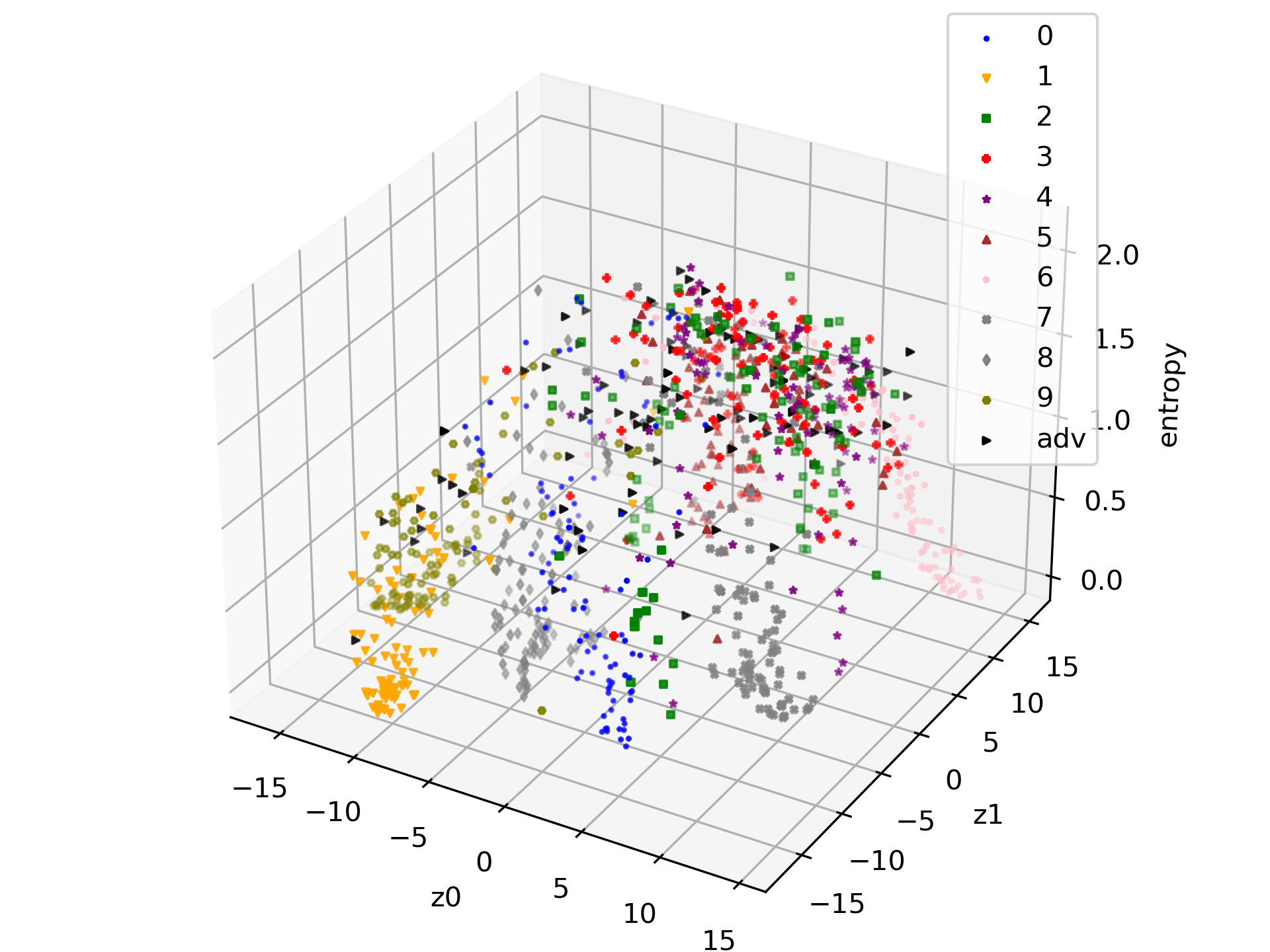}\vspace{-3mm}
\par\end{centering}
}
\par\end{centering}
\caption{t-SNE visualization of the latent space. Experiment on the CIFAR10
dataset with ResNet20 architecture. In each subfigure, the black-triangles
represents for the adversarial examples. The left-subfigure is 2D
t-SNE visualization while the right-subfigure is 2D t-SNE with entropy
of prediction in the z-axis.  \label{fig:tsne}}
\end{figure*}

In addition to the quantitative evaluation as provided in Section
4.4 in the main paper, we provide a qualitative comparison via the
t-SNE visualization as shown in Figure \ref{fig:tsne}. The experiments
have been conducted on the CIFAR10 dataset with ResNet20 architecture
under PGD attack $\{\epsilon=8/255,\eta=2/255,k=250\}$. We visualize
latent representations of 100 adversarial examples in addition to
1000 natural samples of the CIFAR10 dataset. It can be seen that:
(i) In the NAT model as Figure \ref{fig:tsne-NAT}, the latent representations
of natural images are well separate, which explains the high natural
accuracy. However, the adversarial examples also are well separate
and lay on the high confident area of each class (low entropy). It
indicates that, adversarial examples fool the natural model easily
with very high confident. (ii) In the AT model as Figure \ref{fig:tnse-AT},
the latent representations of natural images are less detached, which
explains the lower natural accuracy than the NAT model. The adversarial
examples distribute \emph{randomly} \emph{inside} each cluster.
The predictions of natural images and adversarial examples have higher
entropy which means that the model is less confident. (iii) In our
ASCL and Leaked-ASCL as Figure \ref{fig:tnse-ASCL}, \ref{fig:tnse-LASCL},
the latent representations of natural images are better distinguishable
among classes. More specifically, the adversarial examples' representations
lay in \emph{the boundary} of each cluster, which has higher entropy
than those of natural images.

\section{Additional Experimental Results \label{sec:add-exp-results}}

\subsection{Projection Head in the context of AML \label{sec:project-head}}

\begin{figure}
\begin{centering}
\subfloat[with the projection head\label{fig:with-ph}]{\begin{centering}
\includegraphics[width=0.5\columnwidth]{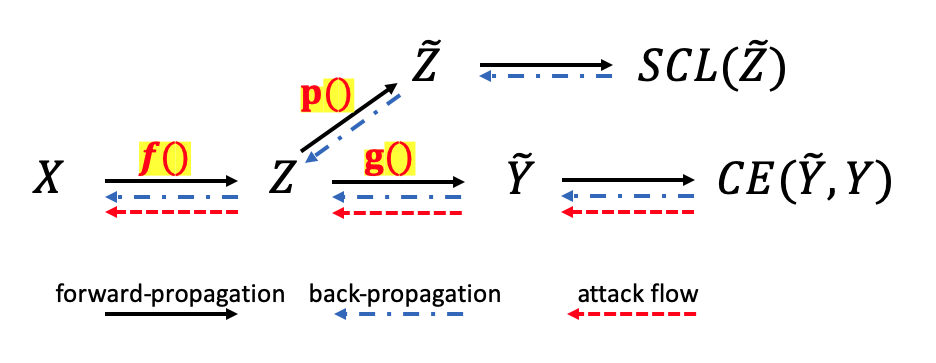}
\par\end{centering}
}
\par\end{centering}
\begin{centering}
\subfloat[without the projection head\label{fig:without-ph}]{\begin{centering}
\includegraphics[width=0.5\columnwidth]{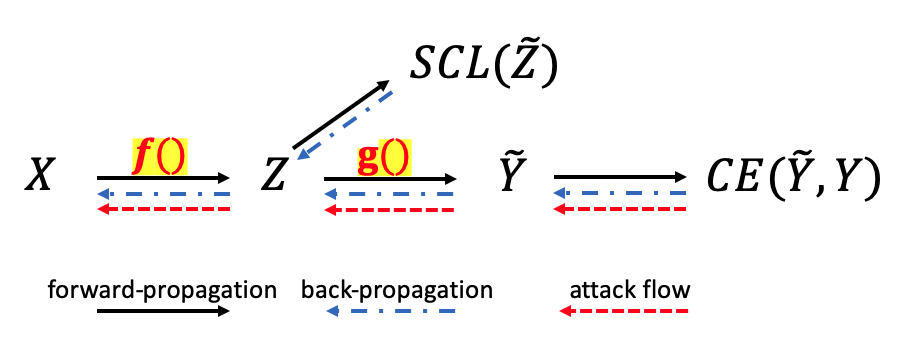}
\par\end{centering}
}
\par\end{centering}
\caption{Training/Attack flows with/without the projection head\label{fig:flow-with-ph}}
\end{figure}

In this section we provide an additional ablation study to further
understand the effect of the projection head in the context of AML.
We apply our methods (ASCL and Leaked ASCL) with three options of
the projection head as shown in Figure \ref{fig:flow-with-ph}: 
\begin{itemize}
    \item A projection head with only single linear layer $\tilde{\bz}=p^{1}(\bz)=W^{1}(\bz)$
    with layer weight $W^{1}\in\mathcal{R}^{h \times \tilde{h}}$, where $h (\tilde{h})$ 
    is the dimensionality of latent $\bz (\tilde{\bz})$. 
    We choose $\tilde{h}=128$ in our experiments. 
    \item A projection head with two fully connected layers without bias $\tilde{\bz}=p^{2}(\bz)=W^{2}\left(Relu(W^{1}(\bz)\right)$
    with layer weight $W^{1}\in\mathcal{R}^{h\times200}$ and $W^{2}\in\mathcal{R}^{200\times128}$
    and 
    \item Identity mapping $\tilde{\bz}=\bz$. 
\end{itemize}

Table \ref{tab:res-with-ph} shows the performances of three options on the CIFAR10 dataset with
ResNet20 architecture.  We observe that the linear projection head
$p^{1}()$ is better than the identity mapping on both natural accuracy
(by around 1\%) and robust accuracy (on average 0.7\%).  
In contrast, the non-linear projection head $p^{2}()$ 
reduces the robust accuracy on average 0.5\%. 

The improvement on the natural accuracy concurs with the finding in \cite{chen2020simple}
which can be explained by the fact that the projection head helps
to reduce the dimensionality to apply the contrastive loss more efficiently.
As shown in Section B.4 in \cite{chen2020simple} that even using
the same output size, the weight of the projection head has relatively
few large eigenvalues, indicating that it is approximately low-rank.

On the other hand, the effect of the projection head to the robust
accuracy is due to its non-linearity. Figure \ref{fig:with-ph} demonstrates
the training flow and attack flow on our framework with the projection
head. The contrastive loss $\mathcal{L}^{SCL}$ is applied in the
projected layer $\tilde{\bz}$ which induces the compactness on the
projected layer but not the intermediate layer $\bz$. Therefore,
when using a non-linear projection head (e.g., $p^{2}$), the compactness
in the intermediate layer is weaker than the projected layer. For
example, a relationship $\left\Vert \tilde{\bz}_{i}-\tilde{\bz}_{j}\right\Vert _{p}\leq\left\Vert \tilde{\bz}_{i}-\tilde{\bz}_{k}\right\Vert _{p}$
in the projected layer can not imply a relationship $\left\Vert \bz_{i}-\bz_{j}\right\Vert _{p}\leq\left\Vert \bz_{i}-\bz_{k}\right\Vert _{p}$
in the intermediate layer. It explains why using the non-linear projection
head reduces the effectiveness of the SCL to the adversarial robustness.

\begin{table}
    \caption{Performance comparison with/without the projection head $p()$ on
    the CIFAR10 dataset with ResNet20 architecture. $p^{1}()$ and $p^{2}()$
    represent for the projection head with one layer and two layers respectively.}
    \label{tab:res-with-ph}
    \begin{centering}
    \resizebox{0.4\textwidth}{!}{\centering\setlength{\tabcolsep}{2pt}
    \begin{tabular}{lcccc}
    \hline 
    & Nat. & PGD & AA\tabularnewline
    \hline 
    ASCL without $p()$ & 76.4 & 52.7 & 40.9\tabularnewline
    ASCL with $p^{1}()$ & 77.3 & \textbf{53.3} & \textbf{41.3}\tabularnewline
    ASCL with $p^{2}()$ & 76.6 & 52.3 & 39.7\tabularnewline
    (Leaked)ASCL without $p()$ & 75.5 & 53.7 & 42.0\tabularnewline
    (Leaked)ASCL with $p^{1}()$ & 76.5 & \textbf{54.1} & \textbf{42.3}\tabularnewline
    (Leaked)ASCL with $p^{2}()$ & 75.7 & 52.9 & 41.1\tabularnewline
    \hline 
    \end{tabular}}
    \par\end{centering}
    \end{table}

\subsection{Contribution of each component in ASCL\label{subsec:Ablation-study}}

We provide an ablation study to investigate the contribution
of each of ASCL's components to the performance and emphasize the importance of our SCL component. 
We experiment on the CIFAR10 dataset with two architectures, i.e., ResNet18 and WideResNet-34-10 (WRN). 
There are two remarks that can be observed from Table \ref{tab:ab-com} such that: 

(i) Using original SCL slightly improves the adversarial robustness against weak adversarial attacks 
but cannot defend strong ones. More specifically, the robust accuracy against PGD with $\epsilon=8/255,k=5,\eta=2/255$
is $1.2\%$ while that for non-defence model is $0.0\%$. 
However, the robust accuracy drops to $0.0\%$ against stronger attacks, i.e., PGD with $\epsilon=8/255,k=250,\eta=2/255$ or Auto-Attack. 
A similar observation was observed in \cite{kim2020adversarial} when the original SCL only achieves $0.08\%$ robust accuracy. 
The result shows that while the original contrastive learning induces weak robustness in DNN models as our analysis in Section 2, 
directly adopting contrastive learning into AML hardly improves the adversarial robustness against strong attacks which emphasizes the importance 
of our adaptions.

(ii) Adding SCL significantly improves the natural performance and adversarial robustness of the model. More specifically, with 
ResNet18 architecture, adding \textit{SCL} to \textit{ADV} can gain improvements of $2.3\%$ of natural accuracy and $1.5\%$ of robust accuracy against Auto-Attack. 
With WideResNet architecture, the improvements of natural/robust accuracies are $2.7\%$ and $1.7\%$, respectively.   
Similar improvements can be observed when adding \textit{SCL} to \textit{ADV+VAT}. 
More specifically, the gaps of natural accuracy with/without \textit{SCL} are $0.8\%$ and $1.1\%$ in experiments with ResNet18 and WideResNet, respectively.
These gaps of robust accuracy against Auto-Attack are $0.3\%$ and $0.7\%$. 

\begin{table}
    \caption{Ablation study on the CIFAR10 dataset with different architectures. \textit{orgSCL} represents 
    the original SCL version with two standard data-augmentations. 
    \textbf{Bold} numbers indicate there are improvements over the previous settings 
    (i.e., only using ADV or using both ADV and VAT) when adding our SCL.}
    \begin{centering}
    \resizebox{0.40\textwidth}{!}{\centering\setlength{\tabcolsep}{2pt}
    \begin{tabular}{llccc}
    \hline 
    & Model & Nat. & PGD & AA\tabularnewline
    \hline 
    orgSCL & ResNet18 & 93.80 & 0.0 & 0.0 \tabularnewline
    ADV & ResNet18 & 82.75 & 52.95 & 48.81\tabularnewline
    ADV+SCL & ResNet18 & \textbf{85.02} & \textbf{53.40} & \textbf{50.31}\tabularnewline
    ADV+VAT & ResNet18 & 83.73 & 53.00 & 49.39\tabularnewline 
    ADV+VAT+SCL & ResNet18 & \textbf{84.54} & \textbf{54.29} & \textbf{49.66}\tabularnewline    
    \hline 
    ADV & WRN & 84.93 & 55.04 & 51.12\tabularnewline
    ADV+SCL & WRN & \textbf{87.70} & 54.05 & \textbf{52.80}\tabularnewline
    ADV+VAT & WRN & 85.96 & 54.76 & 51.87\tabularnewline
    ADV+VAT+SCL & WRN & \textbf{87.12} & \textbf{55.93} & \textbf{52.53}\tabularnewline 
    \hline 
    \end{tabular}}
    \par\end{centering}
    \label{tab:ab-com}
    \end{table}

We provide an additional experiment to further understand the contribution
of each component in our framework. Table \ref{tab:ab-com-more} shows
the result on the CIFAR10 dataset with ResNet20 architecture. We observe
that using SCL alone can helps to improve the natural accuracy, but
enforcing the contrastive loss too much reduces the effectiveness.
On the other hand, increasing the VAT's weight increases the robustness
but significantly reduces the natural performance which concurs with
the finding in \cite{Zhang2019theoretically}. Therefore, to balance
the trade-off between natural accuracy and robustness, we choose $\lambda^{SCL}=1,\lambda^{VAT}=2$
as the default setting in our framework. 


\begin{table}
    \caption{Ablation study with different parameter settings on the CIFAR10 dataset with ResNet20.\label{tab:ab-com-more}}
    \begin{centering}
    \resizebox{0.35\textwidth}{!}{\centering\setlength{\tabcolsep}{2pt}
    \begin{tabular}{lcccc}
    \hline 
    & Nat. & PGD & AA\tabularnewline
    \hline 
    $\lambda^{SCL}=0,\lambda^{VAT}=0$ & 78.8 & 48.1 & 36.1\tabularnewline
    $\lambda^{SCL}=1,\lambda^{VAT}=0$ & 80.1 & 46.5 & 34.7\tabularnewline
    $\lambda^{SCL}=2,\lambda^{VAT}=0$ & 79.5 & 46.7 & 34.7\tabularnewline
    $\lambda^{SCL}=3,\lambda^{VAT}=0$ & 79.6 & 45.8 & 34.4\tabularnewline
    $\lambda^{SCL}=4,\lambda^{VAT}=0$ & 79.2 & 45.6 & 34.3\tabularnewline
    $\lambda^{SCL}=0,\lambda^{VAT}=1$ & 77.4 & 50.6 & 38.2\tabularnewline
    $\lambda^{SCL}=0,\lambda^{VAT}=2$ & 75.4 & 53.0 & 40.0\tabularnewline
    $\lambda^{SCL}=0,\lambda^{VAT}=3$ & 73.3 & 54.4 & 42.3\tabularnewline
    $\lambda^{SCL}=0,\lambda^{VAT}=4$ & 71.2 & 55.0 & 43.1\tabularnewline
    \hline 
    \end{tabular}}
    \par\end{centering}
    \end{table}

\subsection{Global and Local Selections \label{subsec:add-Global-Local}}

\begin{figure}
\begin{centering}
\subfloat[Epoch 1\label{fig:Epoch-1}]{\begin{centering}
\includegraphics[width=0.5\columnwidth]{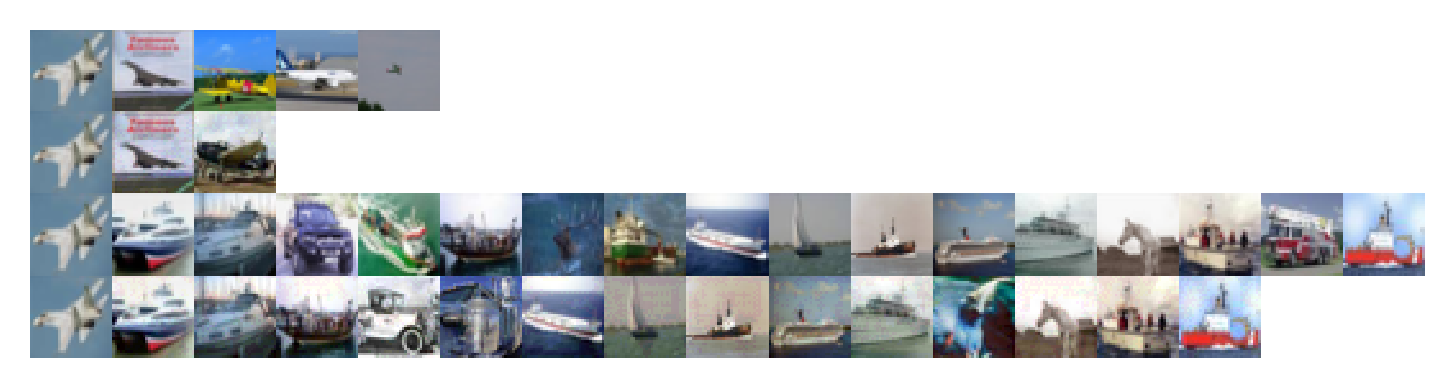}
\par\end{centering}
}
\par\end{centering}
\begin{centering}
\subfloat[Epoch 30\label{fig:Epoch-100}]{\begin{centering}
\includegraphics[width=0.5\columnwidth]{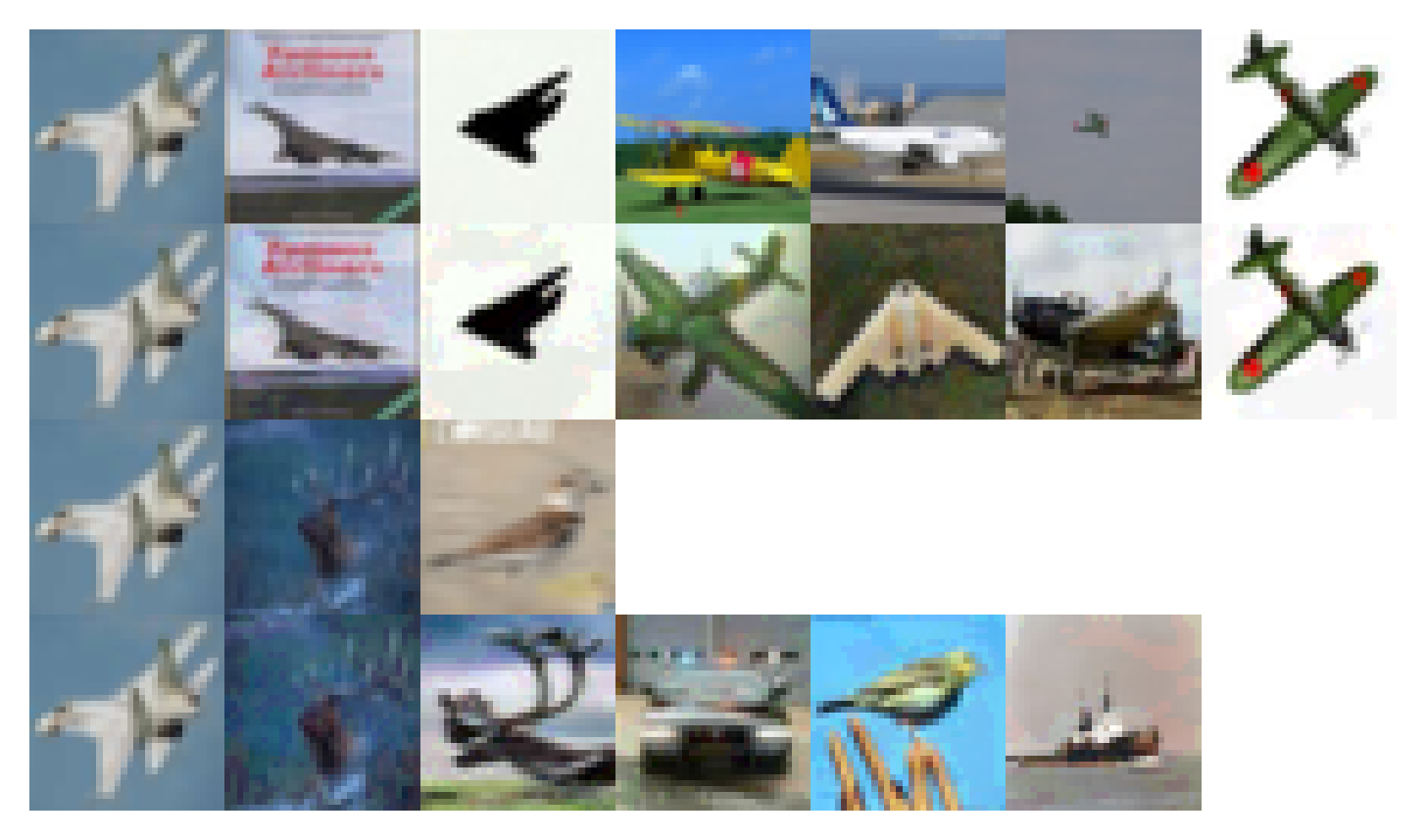}
\par\end{centering}
}
\par\end{centering}
\begin{centering}
\subfloat[Epoch 200\label{fig:Epoch-200}]{\begin{centering}
\includegraphics[width=0.5\columnwidth]{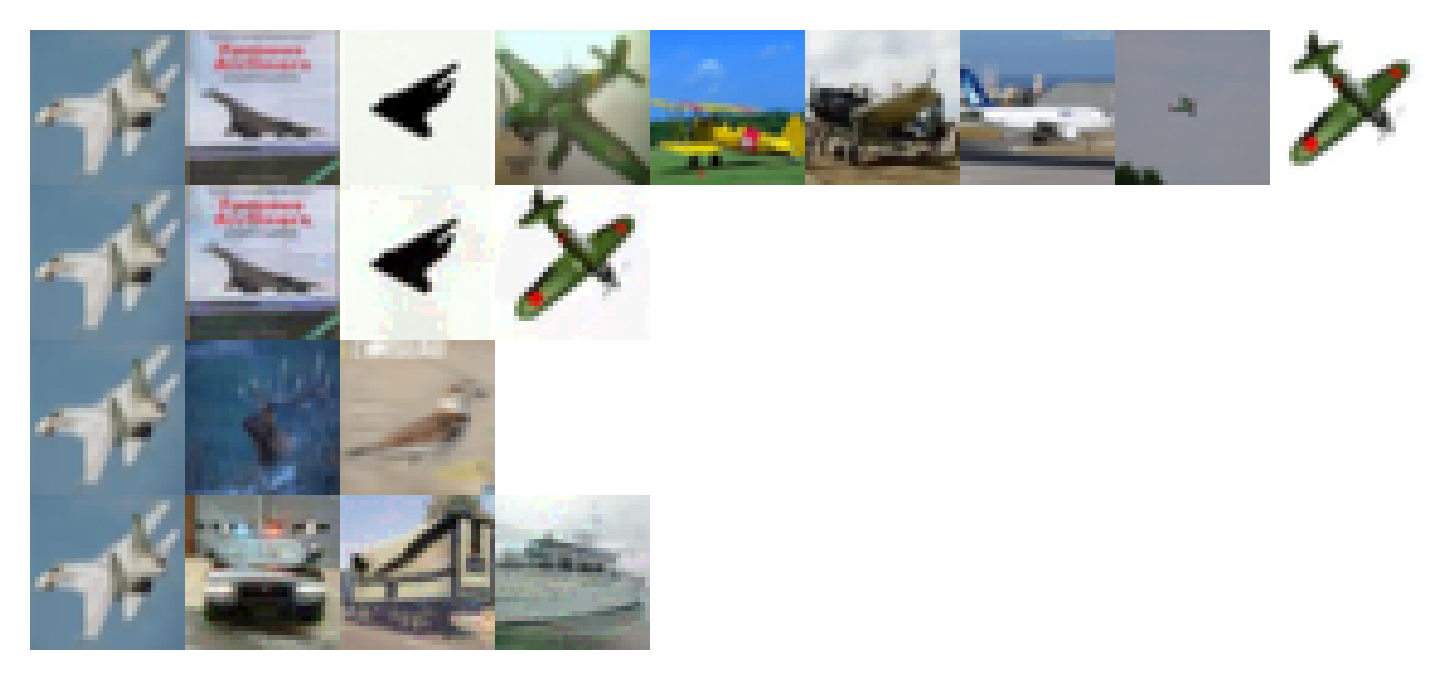}
\par\end{centering}
}
\par\end{centering}
\caption{Positive and negative samples from the Leaked Local Selection strategy.
In each image, the first column represents the anchor followed by
its positive and negative samples. Row 1 and 2 represent the natural
and adversarial positive samples respectively. Row 3 and 4 represent
the natural and adversarial negative samples respectively. \label{fig:pos-neg samples}}
\end{figure}

We provide an example of selected positive and negative samples which
have been chosen by the Leaked-Local Selection as Figure \ref{fig:pos-neg samples}.
It can be seen that, with the same query image, the corresponding
negatives and positives have been selected differently overtime. More
specifically, at the beginning of training progress (epoch 1, Figure
\ref{fig:Epoch-1}), only few positive images (2-4 images) were picked,
while those of negatives are larger (around 14-16 images). Correlating
with the model performance, the number of positive images increases
while the number of negative images decreases. At the end of training
progress (epoch 200, Figure \ref{fig:Epoch-200}) there are 8 natural
images and 3 adversarial images in the positive set, while those in
the negative set are 2 natural images and 3 adversarial images. The
changing of positives/negatives in this example is inline with the
statistic as in Figure 3b in the main paper. In addition, given an
anchor image $\bx_{i}$ , the natural image $\bx_{j}$ and adversarial
image $\bx_{j}^{a}$ ($j\neq i$) have been treated independently
as in Table 1 in the main paper, therefore, we get more flexible in
the positive and negative set, for example, only one of $\bx_{j}$
or $\bx_{j}^{a}$ has been selected as a negative (or a positive)
as in Figure \ref{fig:pos-neg samples}. 

\section{Background and Related works\label{sec:related}}

In this section, we present a fundamental background and related
works to our approach. First, we introduce well-known contrastive learning
frameworks, followed by a brief introduction of adversarial attack
and defense methods. We then provide a comparison of our approach
with defense methods on a latent space, especially, those integrated
with contrastive learning frameworks. 

\subsection{Contrastive Learning}

\subsubsection{General formulation}

Self-Supervised Learning (SSL) became an important tool that helps
Deep Neural Networks exploit structure from gigantic unlabeled data
and transfers it to downstream tasks. The key success factor of SSL
is choosing a pretext task that heuristically introduces interaction
among different parts of the data (e.g., CBOW and Skip-gram \cite{mikolov2013distributed},
predicting rotation \cite{gidaris2018unsupervised}). Recently, Self-Supervised
Contrastive Learning (SSCL) with contrastive learning as the pretext
task surpasses other SSL frameworks and nearly achieves supervised-learning's
performance. The main principle of SSCL is to introduce a contrastive
correlation among visual representations of positives ('similar')
and negatives ('dissimilar') with respect to an anchor one. There
are several SSCL frameworks have been proposed (e.g., MoCo \cite{he2020momentum},
BYOL \cite{grill2020bootstrap}, CURL \cite{srinivas2020curl}), however,
in this section, we mainly introduce the SSCL in \cite{chen2020simple}
which had been integrated with adversarial examples to improve adversarial
robustness in \cite{kim2020adversarial,jiang2020contrastive} followed
by the Supervised Contrastive Learning (SCL) \cite{khosla2020supervised}
which has been used in our approach. 

Consider a batch of N pairs $\{\bx_{i},\by_{i}\}_{i=1}^{N}$ of benign
images and their labels. With two random transformations $\mathcal{T},\mathcal{A}$
we have a set of transformed images $\{\bx_{i}^{\mathcal{T}},\bx_{i}^{\mathcal{A}},\by_{i}\}_{i=1}^{N}$.
The general formulation of contrastive learning as follow: 

\begin{equation}
\mathcal{L}^{\text{CL}}=\frac{1}{N}\sum_{i=1}^{N}\mathcal{L}_{i}^{\mathcal{T},cl}+\mathcal{L}_{i}^{\mathcal{A},cl}\label{eq:general-cl}
\end{equation}

where $\mathcal{L}_{i}^{\mathcal{T}.cl}$ is the contrastive loss
w.r.t. the anchor $\bx_{i}^{\mathcal{T}}$:

\begin{equation}
\mathcal{L}_{i}^{\mathcal{T},cl}=\frac{-1}{\left|\bZ_{i}^{+}\right|+1}\underset{\bz_{j}\in\bZ_{i}^{+}\cup\{\bz_{i}^{\mathcal{A}}\}}{\sum}\log\frac{e^{\frac{sim(\bz_{j},\bz_{i}^{\mathcal{T}})}{\tau}}}{\underset{\bz_{k}\in\bZ_{i}^{+}\cup\bZ_{i}^{-}\cup\{\bz_{i}^{\mathcal{A}}\}}{\sum}e^{\frac{sim(\bz_{k},\bz_{i}^{\mathcal{T}})}{\tau}}}\label{eq:cl-t}
\end{equation}

and $\mathcal{L}_{i}^{\mathcal{A},cl}$ is the contrastive loss w.r.t.
the anchor $\bx_{i}^{\mathcal{A}}$: 

\begin{equation}
\mathcal{L}_{i}^{\mathcal{A},cl}=\frac{-1}{\left|\bZ_{i}^{+}\right|+1}\underset{\bz_{j}\in\bZ_{i}^{+}\cup\{\bz_{i}^{\mathcal{T}}\}}{\sum}\log\frac{e^{\frac{sim(\bz_{j},\bz_{i}^{\mathcal{A}})}{\tau}}}{\underset{\bz_{k}\in\bZ_{i}^{+}\cup\bZ_{i}^{-}\cup\{\bz_{i}^{\mathcal{T}}\}}{\sum}e^{\frac{sim(\bz_{k},\bz_{i}^{\mathcal{A}})}{\tau}}}\label{eq:cl-a}
\end{equation}

The formulation shows the general principle of contrastive learning
such that: (i) $\bZ_{i}^{+}\cup\bZ_{i}^{-}\cup\{\bz_{i}^{\mathcal{A}},\bz_{i}^{\mathcal{T}}\}=\{\bz_{j}^{\mathcal{T}},\bz_{j}^{\mathcal{A}}\}_{j=1}^{N}\;\forall i\in[1,N]$
where $\bZ_{i}^{+}$ and $\bZ_{i}^{-}$ are positive and negative
sets which are defined differently depending on self-supervised/supervised
setting, (ii) without loss of generality, in Equation \ref{eq:cl-t},
the similarity $e^{\frac{sim(\bz_{j},\bz_{i}^{\mathcal{T}})}{\tau}}$
between the anchor $\bz_{i}^{\mathcal{T}}$ and a positive sample
$\bz_{j}\in\bZ_{i}^{+}\cup\{\bz_{i}^{\mathcal{A}}\}$ has been normalized
with sum of all possible pairs between the anchor and the union set
of $\bZ_{i}^{+}\cup\bZ_{i}^{-}\cup\{\bz_{i}^{\mathcal{A}}\}$ to ensures
that the log argument is not higher than 1, (iii) the contrastive
loss in Equation \ref{eq:cl-t} pulls anchor representation $\bz_{i}^{\mathcal{T}}$
and the positives' representations $\bZ_{i}^{+}\cup\{\bz_{i}^{\mathcal{A}}\}$
close together while pushes apart those of negatives $\bZ_{i}^{-}$.

\paragraph{Explanation for our Formulation.}
It is worth noting that, our derivation shows the general formulation
of the contrastive learning which can be adapted to SSCL \cite{chen2020simple},
SCL \cite{khosla2020supervised} or our Local ASCL by defining the
positive and negative sets differently. Moreover, by using terminologies
positive set $\bZ_{i}^{+}$ and those sample from the same instance
$\{\bz_{i}^{\mathcal{T}},\bz_{i}^{\mathcal{A}}\}$ separately, we
emphasize the importance of the anchor's transformation which stand
out other positives. Last but not least, our derivation normalizes
the contrastive loss in Equation \ref{eq:general-cl} to the same
scale with the cross-entropy loss and the VAT loss as in Section 3,
which helps to put all terms together appropriately. 

\paragraph{Self-Supervised Contrastive Learning.}

In SSCL\cite{chen2020simple}, the positive set (excluding those samples
from the same instance $\bz_{i}^{\mathcal{A}},\bz_{i}^{\mathcal{T}}$)
$\bZ_{i}^{+}=\emptyset$ $(\left|\bZ_{i}^{+}\right|=0)$ while the
negative set $\bZ_{i}^{-}=\{\bz_{j}^{\mathcal{T}},\bz_{j}^{\mathcal{A}}\mid j\neq i\}$
which includes all other samples except those from the same instance
$\bz_{i}^{\mathcal{A}},\bz_{i}^{\mathcal{T}}$. In this case, the
formulation of SSCL as follow: 

\begin{equation}
\mathcal{L}_{i}^{\mathcal{T},sscl}=-\log\frac{e^{\frac{sim(\bz_{j},\bz_{i}^{\mathcal{T}})}{\tau}}}{\underset{\bz_{k}\in\bZ_{i}^{-}\cup\{\bz_{i}^{\mathcal{A}}\}}{\sum}e^{\frac{sim(\bz_{k},\bz_{i}^{\mathcal{T}})}{\tau}}}
\end{equation}

and 

\begin{equation}
\mathcal{L}_{i}^{\mathcal{A},sscl}=-\log\frac{e^{\frac{sim(\bz_{j},\bz_{i}^{\mathcal{A}})}{\tau}}}{\underset{\bz_{k}\in\bZ_{i}^{-}\cup\{\bz_{i}^{\mathcal{T}}\}}{\sum}e^{\frac{sim(\bz_{k},\bz_{i}^{\mathcal{A}})}{\tau}}}
\end{equation}

\paragraph{Supervised Contrastive Learning.}

The SCL framework leverages the idea of contrastive learning with
the presence of label supervision to improve the regular cross-entropy
loss. The positive set and the negative set are $\bZ_{i}^{+}=\{\bz_{j}^{\mathcal{T}},\bz_{j}^{\mathcal{A}}\mid j\neq i,\by_{j}=\by_{i}\}$
and $\bZ_{i}^{-}=\{\bz_{j}^{\mathcal{T}},\bz_{j}^{\mathcal{A}}\mid j\neq i,\by_{j}\neq\by_{i}\}$,
respectively. As mentioned in \cite{khosla2020supervised}, there
is a major advantage of SCL compared with SSCL in the context of regular
machine learning. Unlike SSCL in which each anchor has only single
positive sample, SCL takes advantages of the labels to have many positives
in the same batch size N. This strategy helps to reduce the false
negative cases in SSCL when two samples in the same class are pushed
apart. As shown in \cite{khosla2020supervised}, the SCL training
is more stable than SSCL and also achieves a better performance.

\subsubsection{Important factors for Contrastive Learning}

\paragraph{Data augmentation.}

Chen et al. \cite{chen2020simple} empirically found that SSCL needs
stronger data augmentation than supervised learning. While the SSCL's
performance experienced a huge gap of 5\% with different data augmentation
(Table 1 in \cite{chen2020simple}), the supervised performance was
not changed much with the same set of augmentation. Therefore, in
our paper, to reduce the space of hyper-parameters we use only one
adversarial transformation $\mathcal{A}$(e.g., PGD\cite{madry2017towards}
or TRADES\cite{Zhang2019theoretically}) while using the identity
transformation $\mathcal{T}$, $\bx_{i}^{\mathcal{T}}=\bx_{i}$ ($\bz_{i}^{\mathcal{T}}=\bz_{i}$),
and let the investigation of using different data augmentations for
future works. 

\paragraph{Batch size.}

As shown in Figure 9 in \cite{chen2020simple}, the batch size is
an important factor that strongly affects the performance of the contrastive
learning framework. A larger batch size comes with larger positive
and negative sets, which helps to generalize the contrastive correlation
better and therefore improves the performance. He et al. \cite{he2020momentum}
proposed a memory bank to store the previous batch information which
can lessen the batch size issue. In our framework, because of the
limitation on computational resources, we only tried with a small
batch size (128) which likely limits the contribution of our methods.

\paragraph{Projection head.}

Normally, the representation vector which is the output of the encoder
network has very high dimensionality, e.g., the final pooling layer
in ResNet-50 and ResNet-200 has 2048 dimensions. Therefore, applying
contrastive learning directly on this intermediate layer is less effective.
Alternatively, CL frameworks usually use a projection network $p()$
to project the normalized representation vector $\bz$ into a lower
dimensional vector $\tilde{\bz}=p(\bz)$ which is more suitable for
computing the contrastive loss. To avoid over-parameterized, CL frameworks
usually choose a small projection head with only one or two fully-connected
layers.  

\subsection{Adversarial attack}

\paragraph*{Projected Gradient Decent (PGD).}

is an iterative version of the FGSM attack \cite{goodfellow2014explaining}
with random initialization \cite{madry2017towards}. It first randomly
initializes an adversarial example in a perturbation ball by adding
uniform noise to a clean image, followed by multiple steps of one-step
gradient ascent, at each step projecting onto the perturbation ball.
The formula for the one-step update is as follows: 
\begin{equation}
x_{a}^{t+1}=\text{Proj\ensuremath{_{B_{\varepsilon}\left(x\right)}}}(x_{a}^{t}+\eta\;\text{sign}\left(\nabla\ell(x,y,\theta)\right)
\end{equation}
where $B_{\varepsilon}\left(x\right)\triangleq\left\{ x':\norm{x'-x}<\varepsilon\right\} $
is the perturbation ball with radius $\varepsilon$ around $x$and
$\eta$ is the gradient scale for each step update.

\paragraph*{Auto-Attack. }

Even the most popular attack, PGD can still fail in some extreme
cases \cite{croce2019scaling} because of two issues: (i) fixed step
size $\eta$ which leads to sub-optimal solutions and (ii) the sensitivity
of a gradient to the scale of logits in the standard cross-entropy
loss. Auto-Attack \cite{croce2020reliable} proposed two variants
of PGD to deal with these potential issues by (i) automatically selecting
the step size across iterations (ii) an alternative logit loss which
is both shift and rescaling invariant. Moreover, to increase the diversity
among the attacks used, Auto-Attack combines two new versions of PGD
with the white-box attack FAB \cite{croce2019minimally}and the blackbox
attack Square Attack \cite{andriushchenko2020square} to form a parameter-free,
computationally affordable, and user-independent ensemble of complementary
attacks to estimate adversarial robustness. Therefore, besides PGD,
Auto-Attack is considered as the new standard evaluation for adversarial
robustness. 

\subsection{Adversarial defense }

\subsubsection{Adversarial training}

Adversarial training (AT) originate in \cite{goodfellow2014explaining},
which proposed incorporating a model's adversarial examples into training
data to make the model's loss surface to be smoother, thus, improve
its robustness. Despite its simplicity, AT \cite{madry2017towards}
was among the few that were resilient against attacks other than gave
a false sense of robustness because of the obfuscated gradient \cite{athalye2018obfuscated}.
To continue its success, many AT's variants have been proposed including
(1) different types of adversarial examples (e.g., the worst-case
examples \cite{goodfellow2014explaining} or most divergent examples
\cite{Zhang2019theoretically}), (2) different searching strategies
(e.g., non-iterative FGSM, Rand FGSM with a random initial point or
PGD with multiple iterative gradient descent steps \cite{madry2017towards}),
(3) additional regularizations, e.g., adding constraints in the latent
space \cite{zhang2019defense,bui2020improving}, (4) difference in
model architecture, e.g., activation function \cite{xie2020smooth}
or ensemble models \cite{pang2019improving}.

\subsubsection{Defense with a latent space }

Unlike an input space $\bX$, a latent space $\bZ$ has a lower dimensionality
and a higher mutual information with the prediction space than the
input one $I(Z,Y)\geq I(X,Y)$ \cite{tishby2015deep}. Therefore,
defense with the latent space has particular characteristics to deal
with adversarial attacks notably \cite{zhang2019defense,bui2020improving,mao2019metric,xie2019feature,samangouei2018defense}.
For example, DefenseGAN \cite{samangouei2018defense} used a pretrained
GAN which emulates the data distribution to generate a denoised version
of an adversarial example. On the other hand, instead of removing
noise in the input image, Xie et al. \cite{xie2019feature} attempted
to remove noise in the feature space by using non-local means as a
denoising block. However, these works were criticized by \cite{athalye2018obfuscated}
as being easy to attack by approximating the backward gradient signal.

\subsubsection{Defense with contrastive learning }

The idea of defense with contrastive correlation in the latent space
can be traced back to \cite{mao2019metric} which proposed an additional
triplet regularization to adversarial training. However, the triplet
loss can only handle one positive and negative at a time, moreover,
requires computationally expensive hard negative mining \cite{schroff2015facenet}.
As discussed in \cite{khosla2020supervised}, the triplet loss is
a special case of the contrastive loss when the number of positives
and negatives are each one and has lower performance in general than
the contrastive loss. Recently, \cite{jiang2020contrastive,kim2020adversarial}
integrated SSCL \cite{chen2020simple} to learn unsupervised robust
representations for improving robustness in unsupervised/semi-supervised
setting. Specifically, both methods proposed a new kind of adversarial
examples which is based on the SSCL loss instead of regular cross-entropy
loss \cite{goodfellow2014explaining} or KL divergence \cite{Zhang2019theoretically}.
By adversarially pre-training with these adversarial examples, the
encoder is robust against the instance-wise attack and obtains comparable
robustness to supervised adversarial training as reported in \cite{kim2020adversarial}.
On the other hand, Jiang et al. \cite{jiang2020contrastive} proposed
three options of pre-training. However, their best method made use
of two adversarial examples that requires a much higher computational
cost to generate. Although these above works have the similar general
idea of using contrastive learning to improve adversarial robustness
with ours, we choose to compare our methods with RoCL-AT/TRADES in \cite{kim2020adversarial}
which is most close to our problem setting. More specifically, after
pre-training phase with adversarial examples w.r.t. the contrastive
loss, RoCL-AT/TRADES apply fine-tuning with standard supervised adversarial
training, which requires full label. We use the reported result as
in Table 1 in \cite{kim2020adversarial} which used a larger batch
size (256). It is a worth noting that the best reported version RoCL-AT-SS achieves 
91.34\% natural accuracy and 49.66\% robust accuracy is a fine-tuned on 
a ImageNet pretrained model with self-supervised loss (e.g., SimCLR \cite{chen2020simple}), 
therefore, is not as a reference for comparison.

Most closely related to our work is \cite{bui2020improving} which
also aims to realize the compactness in latent space to improve the
robustness in supervised setting. They proposed a label weighting
technique that sets the positive weight to the divergence of two examples
in the same class and negative weight in any other cases. Therefore,
when minimizing the divergence loss with label weighting, the divergences
of those in the same class (positives) are encouraged to be close
together, while those of different classes (negatives) to be distant.

\end{document}